\documentclass{article}

\usepackage[preprint]{neurips_2026}

\usepackage[utf8]{inputenc} 
\usepackage[T1]{fontenc}    
\usepackage{hyperref}       
\usepackage{url}            
\usepackage{booktabs}       
\usepackage{amsfonts}       
\usepackage{nicefrac}       
\usepackage{microtype}      
\usepackage{xcolor}         
\usepackage{amsmath}
\usepackage{tabularx}
\usepackage{graphicx}
\usepackage{algorithm}
\usepackage{algpseudocode}
\usepackage{float}

\newcommand{\wrap}{\operatorname{wrap}}
\newcommand{\diag}{\operatorname{diag}}

\newcommand{\Cat}{\operatorname{Categorical}}

\title{Learning to Bias: Machine Learning-Enhanced Particle Filters}

\author{%
  Apoorv Srivastava\thanks{Corresponding author} \\
  Stanford University\\
  Stanford, CA 94305 \\
  \texttt{apoorv1@stanford.edu} \\
  \And
  Eric Darve \\
  Stanford University\\
  Stanford, CA 94305 \\
  \texttt{darve@stanford.edu} \\
}

\begin{document}

\maketitle

\begin{abstract}
    Sequential inference estimates latent states from noisy and incomplete observations. Particle Filters (PFs), a class of Monte Carlo methods based on importance sampling, provide a flexible framework for this task, but often suffer from poor sample efficiency and unfavorable scaling with dimension, partly due to suboptimal proposal distributions. We address these challenges by integrating learned proposals into the PF framework. We introduce Neural Optimal Particle Filters (NOPFs), which learn an amortized approximation to the optimal proposal from offline simulated one-step conditioning tuples. The learned proposal is used as a drop-in replacement in standard PF updates, with samples corrected by standard importance weights so that the method asymptotically targets the same filtering distribution under standard support and density-evaluation assumptions. Across stochastic nonlinear benchmarks of varying inference complexity, NOPFs improve sample efficiency and distributional accuracy over standard PF baselines with modest computational overhead. The approach integrates data-driven proposal learning into classical inference without altering the underlying filtering objective.
\end{abstract}

\section{Introduction}

Sequential inference aims to estimate latent states from noisy and incomplete observations and is central to applications such as robotics, tracking, weather forecasting, and data assimilation~\cite{Wills2023, Fearnhead2018}. Particle Filters (PFs), also known as sequential Monte Carlo (SMC) methods, provide a flexible Monte Carlo approximation of the posterior distribution over the current state using weighted particles.

PFs rely on importance sampling, and their sample efficiency depends critically on the proposal distribution. The number of particles required can grow sharply with state dimension, leading to poor scaling in high-dimensional settings~\cite{vanLeeuwen2019}. Naive proposals often place particles in regions unlikely under the observations, so most particles receive negligible weights and contribute little to the posterior estimate. Effective proposals should therefore use the current observation to guide particles toward plausible posterior regions while still allowing proper importance weighting.

In this work, we introduce Neural Optimal Particle Filters (NOPFs), which learn proposal distributions that serve as plug-and-play replacements in standard PF updates. Samples from the learned proposal are corrected using the standard importance-weighting update, thereby preserving the underlying filtering objective while improving particle placement.

This work makes three contributions. First, we formulate proposal learning for known state-space models as an offline amortized approximation to the locally optimal one-step proposal, rather than as end-to-end model learning or trajectory-level SMC adaptation. Second, we introduce a residual Gaussian proposal around the deterministic transition skeleton, enabling efficient sampling and tractable density evaluation for valid SIS weight correction. Third, we evaluate NOPFs on nonlinear localization benchmarks spanning approximately Gaussian, multimodal, and data-association regimes, comparing against Kalman-based and proposal-enhanced PF baselines.

\section{Background}
\label{sec:background}

Given a latent state-space model with state $\mathbf{x}_t$ and observation $\mathbf{y}_t$ at time $t$, the system evolves according to
\begin{equation}
    \mathbf{x}_{t} = \mathbf{f}(\mathbf{x}_{t-1}, \mathbf{v}_t), 
    \qquad 
    \mathbf{y}_t = \mathbf{h}(\mathbf{x}_t, \mathbf{w}_t),
    \label{eqn:problem_setup}
\end{equation}
where $\mathbf{v}_t$ and $\mathbf{w}_t$ denote process and observation noise. The transition model $\mathbf{f}$ and observation model $\mathbf{h}$ induce the transition density $p(\mathbf{x}_t \mid \mathbf{x}_{t-1})$ and likelihood $p(\mathbf{y}_t \mid \mathbf{x}_t)$, respectively. The objective is to recursively approximate the filtering distribution $p(\mathbf{x}_t \mid \mathbf{y}_{1:t})$ using the previous posterior $p(\mathbf{x}_{t-1} \mid \mathbf{y}_{1:t-1})$ and the new observation $\mathbf{y}_t$.

For linear-Gaussian models, the Kalman filter provides an analytic solution, with variants such as the EKF, UKF, and EnKF extending this idea to nonlinear settings under approximate Gaussian assumptions~\cite{Evensen2009, wan2000unscented, Gustafsson2018}. PFs, in contrast, represent the filtering distribution nonparametrically as
\begin{equation}
    p(\mathbf{x}_t \mid \mathbf{y}_{1:t})
    \approx
    \sum_{i=1}^N w_t^{(i)}
    \delta(\mathbf{x}_t - \mathbf{x}_t^{(i)}),
    \label{eqn:particle_approximation}
\end{equation}
where $\{\mathbf{x}_t^{(i)}, w_t^{(i)}\}_{i=1}^N$ are particles with non-negative normalized weights satisfying $\sum_{i=1}^N w_t^{(i)}=1$.

PFs approximate the filtering update using Sequential Importance Sampling (SIS). At time $t$, each particle is sampled from a proposal, $\mathbf{x}_t^{(i)} \sim q(\mathbf{x}_t \mid \mathbf{x}_{t-1}^{(i)}, \mathbf{y}_t)$, and assigned the importance weight
\begin{equation}
    w_t^{(i)}
    \propto
    w_{t-1}^{(i)}
    \frac{
    p(\mathbf{y}_t \mid \mathbf{x}_t^{(i)})
    p(\mathbf{x}_t^{(i)} \mid \mathbf{x}_{t-1}^{(i)})
    }{
    q(\mathbf{x}_t^{(i)} \mid \mathbf{x}_{t-1}^{(i)}, \mathbf{y}_t)
    },
    \label{eqn:importance_sampling}
\end{equation}
followed by normalization. This update follows from the standard SIS factorization~\cite{Doucet2000, Arulampalam2002}. The normalized weights determine each particle's contribution to the posterior approximation. When the proposal places many particles in regions with low likelihood under the new observation, most weights become negligible, and only a few particles dominate the approximation. This weight degeneracy reduces the effective number of useful particles, even when the nominal ensemble size is large. A standard way to stabilize PFs is to resample after normalization.

Resampling mitigates weight degeneracy by duplicating high-weight particles, discarding low-weight particles, and resetting weights to $1/N$. However, resampling acts only after the particles have already been proposed and weighted. If most particles are proposed in low-likelihood regions, it can only replicate the few survivors, reducing diversity and causing sample impoverishment~\cite{Doucet2000, Arulampalam2002}. Thus, proposal design remains central to PF efficiency.

For a fixed ancestor particle $\mathbf{x}_{t-1}$ and new observation $\mathbf{y}_t$, the locally optimal proposal is
\begin{align}
    q^\star(\mathbf{x}_t \mid \mathbf{x}_{t-1}, \mathbf{y}_t) = p(\mathbf{x}_t \mid \mathbf{x}_{t-1}, \mathbf{y}_t).
    \label{eqn:optimal_proposal}
\end{align}
The term local reflects the recursive SIS setting in which $\mathbf{y}_t$ is used to sample only the new state $\mathbf{x}_t$ and does not alter the ancestor locations $\mathbf{x}_{t-1}^{(i)}$. Within this class, the optimal proposal gives the least-degenerate benchmark~\cite{Snyder2015}. It also minimizes the conditional variance of the incremental importance weight given the ancestor particle and current observation~\cite{Doucet2000}. Under this choice, $w_t^{(i)} \propto w_{t-1}^{(i)}p(\mathbf{y}_t \mid \mathbf{x}_{t-1}^{(i)})$, so the updated weight is independent of the newly sampled state $\mathbf{x}_t^{(i)}$. It therefore provides a natural target for learned proposal models. By approximating $p(\mathbf{x}_t \mid \mathbf{x}_{t-1}, \mathbf{y}_t)$, the learned proposal biases particles toward dynamically plausible and observation-consistent regions while preserving valid importance weighting.

In practice, the locally optimal proposal is rarely available in closed form. The Bootstrap Particle Filter (BPF) uses the transition density $p(\mathbf{x}_t \mid \mathbf{x}_{t-1})$ as the proposal, which is simple but ignores the current observation during sampling. Several PF variants, therefore, construct more informative proposals or alter the sampling procedure. Auxiliary PFs use predictive-likelihood approximations to select promising ancestors~\cite{Pitt1999}, Unscented PFs build local Gaussian proposals using unscented transforms~\cite{van2000unscented}, Implicit PFs move particles through particle-wise optimization~\cite{Chorin2010}, and Nudged PFs use likelihood gradients to push particles toward high-likelihood regions~\cite{Akyildiz2019}. More recent score-based filters use learned score models to generate posterior-aligned samples~\cite{Bao2024}.

A related line of work uses SMC not only for filtering with known models, but also within broader parameter-learning and joint model-state inference frameworks. Nested particle filters estimate static model parameters together with latent states using a two-layer particle approximation~\cite{Crisan2018}. Filtering Variational Objectives~\cite{Maddison2017fvo}, Variational Sequential Monte Carlo~\cite{Naesseth2018vsmc}, and Auto-Encoding Sequential Monte Carlo~\cite{anh2018autoencoding} use SMC-based variational objectives or likelihood estimators to learn sequential latent-variable models and associated proposal distributions. Differentiable particle filtering methods modify resampling or the PF computation graph so that model components, such as the forward model and observation operator, can be trained end-to-end~\cite{DifferentiableParticleFiltering, Jonschkowski2018, Karkus2018}, while adaptive SMC methods tune proposal mechanisms using online criteria~\cite{Cornebise2008}. More broadly, learned generative models have also been used to construct importance densities for complex target distributions~\cite{Mller2019}.

Several prior methods are closer to learned or approximate proposal construction. Inference networks for SMC learn amortized stochastic inverses for directed graphical models and use them as proposal distributions~\cite{Paige2016InferenceNetworks}. Computational Doob $h$-transforms approximate fully adapted proposals for discretely observed diffusions using neural approximations of backward Kolmogorov equations~\cite{Chopin2023Doob}. Neural Adaptive Sequential Monte Carlo (NASMC) learns neural proposal distributions within an adaptive SMC framework~\cite{gu2015neuraladaptive}. NASMC is closest in spirit to the present work, but its learning problem is different. It optimizes an inclusive trajectory-level KL objective, with gradients typically estimated from SMC particles, weights, and ancestry variables, and uses proposal models that may depend on latent and observation histories.

In contrast, this work assumes known transition and observation models and focuses specifically on learning an amortized approximation to the one-step Markov proposal $p(\mathbf{x}_t \mid \mathbf{x}_{t-1},\mathbf{y}_t)$. NOPF trains this proposal offline from simulated one-step conditioning tuples, using an ELBO based on analytic model densities, parameterizes a residual Gaussian proposal using the deterministic transition skeleton, and inserts the learned proposal directly into the standard PF update. At inference time, particles sampled from the learned proposal are corrected using the standard importance-weight update in Eq.~\ref{eqn:importance_sampling}, so no online proposal optimization, model learning, differentiable resampling, or weighted SMC training trajectories are required.

\section{Method}
\label{sec:method}
The standard PF update at time $t$ is shown in Alg.~\ref{alg:pf}. Different proposal choices yield different PF variants discussed in Section~\ref{sec:background}. Although richer proposal classes may depend on the particle index $i$, particle histories $\mathbf{x}_{0:t-1}^{(i)}$, or the full previous weighted ensemble $\{(\mathbf{x}_{t-1}^{(j)},w_{t-1}^{(j)})\}_{j=1}^N$, the Markov structure implies that the locally optimal one-step proposal for the $i$th particle reduces to $p(\mathbf{x}_t \mid \mathbf{x}_{t-1}^{(i)}, \mathbf{y}_t)$~\cite{Doucet2000,vanLeeuwen2019}. We therefore learn a conditional density $q_\theta(\mathbf{x}_t \mid \mathbf{x}_{t-1}^{(i)}, \mathbf{y}_t, t)$ to approximate this local proposal, yielding what we call the Neural Optimal Particle Filter (NOPF).

\begin{algorithm}[H]
\caption{Standard Particle Filter Update (Time $t$)}
\label{alg:pf}
\footnotesize
\begin{algorithmic}[1]
\State \textbf{Sample:} Draw $x_t^{(i)} \sim q(x_t \mid x_{t-1}^{(i)}, y_t)$ for $i=1,\ldots,N$.
\State \textbf{Weight:} Update weight using Eq.~\ref{eqn:importance_sampling} then normalize to get $\{w_t^{(i)}\}_{i=1}^N$ such that $\sum_{i=1}^N w_t^{(i)} = 1$.
\State \textbf{Resample:} Resample using weight $w_t^{(i)}$ and reset $w_t^{(i)} = 1/N$.
\end{algorithmic}
\end{algorithm}

\subsection{Learned Proposal Density}

To support importance sampling, the proposal model must allow both efficient sampling and tractable density evaluation. This excludes implicit generative models whose densities are unavailable, and motivates tractable conditional density models such as Gaussian distributions, mixtures of Gaussians, or normalizing flows. Our objective is to learn a family of distributions parameterized by $(\mathbf{x}_{t-1}, \mathbf{y}_t, t)$, such that $q_\theta(\mathbf{x}_t \mid \mathbf{x}_{t-1}, \mathbf{y}_t) \approx p(\mathbf{x}_t \mid \mathbf{x}_{t-1}, \mathbf{y}_t)$. Time $t$ is also provided as an input to the proposal network in all implementations, but we suppress this dependence in the notation for readability.

Although expressive models such as normalizing flows could be used, we adopt a simpler Gaussian proposal. For a fixed ancestor particle $\mathbf{x}_{t-1}^{(i)}$, the local proposal only needs to describe the region reachable under the transition density, reweighted by the current likelihood. While this conditional proposal can in principle be multimodal due to ambiguities in the observation likelihood, in many practical settings such as robotics, the transition density $p(\mathbf{x}_t \mid \mathbf{x}_{t-1})$ is typically unimodal and concentrated around deterministic dynamics~\cite[Chapter 5]{10.5555/1121596}, which often induces a dominant mode in the conditional proposal. The full filtering distribution can still be multimodal, since it is represented by a weighted mixture over many particles. We therefore parameterize the per-particle proposal as a Gaussian density,
\begin{align}
    q_\theta(\mathbf{x}_t \mid \mathbf{x}_{t-1}, \mathbf{y}_t)
    =
    \mathcal{N}\! \left(
    \mathbf{x}_t;
    \boldsymbol{\mu}_{\theta_1}(\mathbf{x}_{t-1}, \mathbf{y}_t, t),
    \boldsymbol{\Sigma}_{\theta_2}(\mathbf{x}_{t-1}, \mathbf{y}_t, t)
    \right),
\end{align}
where $\boldsymbol{\mu}_{\theta_1}$ and $\boldsymbol{\Sigma}_{\theta_2}$ are neural networks, and $\theta=(\theta_1,\theta_2)$ denotes the trainable parameters. This parameterization makes sampling and density evaluation inexpensive, requiring only a forward pass through the network and a Gaussian density evaluation.

Directly learning the map from $(\mathbf{x}_{t-1}^{(i)}, \mathbf{y}_t, t)$ to the proposal parameters would require the network to also learn the deterministic component of the transition model. To reduce this burden, we first propagate each particle through the deterministic skeleton of the dynamics to obtain a nominal prediction $\hat{\mathbf{x}}_t^{(i)}$\footnote{For example, $\hat{\mathbf{x}}_t^{(i)}$ may be chosen as $\mathbb{E}[\mathbf{x}_t \mid \mathbf{x}_{t-1}^{(i)}]$ or $\arg\max_{\mathbf{x}_t} p(\mathbf{x}_t \mid \mathbf{x}_{t-1}^{(i)})$. This deterministic prediction is analogous to the summary statistic used in Auxiliary Particle Filters to evaluate predictive likelihoods.}. The neural networks then use $\hat{\mathbf{x}}_t^{(i)}$, $\mathbf{y}_t$, and $t$ to output the mean and covariance of the proposal. This lets the model learn an observation-informed correction around the predicted state rather than relearning the forward dynamics. Details of the mean and covariance parameterization and network architecture are provided in Appendix~\ref{adx:proposal_parameterization}.

\subsection{Training and Inference}

Rather than learning a single proposal distribution, NOPF learns a family of conditional proposals indexed by the conditioning tuple $(\mathbf{x}_{t-1},\mathbf{y}_t,t)$. The proposal is trained offline by minimizing the expected KL divergence
\begin{align}
    \mathbb E_{(\mathbf x_{t-1},\mathbf y_t,t)} D_{\mathrm{KL}}\!\left(q_\theta(\mathbf x_t \mid \mathbf x_{t-1},\mathbf y_t) \,\|\,  p(\mathbf x_t \mid \mathbf x_{t-1},\mathbf y_t) \right),
\label{eqn:expected_kl_div}
\end{align}
where the expectation is taken over conditioning tuples generated from simulated trajectories.

For a single tuple $(\mathbf x_{t-1},\mathbf y_t,t)$, the ELBO identity derived in Appendix~\ref{adx:loss_function} gives the loss
\begin{align}
\label{eqn:l_elbo}
    \mathcal L_{\mathrm{ELBO}}(\theta;\mathbf x_{t-1},\mathbf y_t,t) = -\mathbb E_{q_\theta} \left[ \log p(\mathbf y_t \mid \mathbf x_t) + \log p(\mathbf x_t \mid \mathbf x_{t-1}) - \log q_\theta(\mathbf x_t \mid \mathbf x_{t-1},\mathbf y_t) \right],
\end{align}
which is equivalent, up to an additive constant independent of $\theta$, to the KL divergence in Eq.~\ref{eqn:expected_kl_div}. The objective minimized in practice is therefore
\begin{align}
    \mathcal L(\theta) = \mathbb E_{(\mathbf x_{t-1},\mathbf y_t,t)} \left[ \mathcal L_{\mathrm{ELBO}}(\theta;\mathbf x_{t-1},\mathbf y_t,t) \right].
\end{align}
This objective encourages the proposal to place probability mass on states that both explain the current observation and remain likely under the transition dynamics, while penalizing mismatch with the proposal density.

Training tuples $(\mathbf{x}_{t-1},\mathbf{y}_t,t)$ are generated by simulating trajectories from the known forward model. This decouples proposal learning from the online filtering procedure, in contrast to learning-based and differentiable PF methods that train through SMC rollouts, weighted particle trajectories, or resampling operations~\cite{Maddison2017fvo, Naesseth2018vsmc, anh2018autoencoding, DifferentiableParticleFiltering, Jonschkowski2018, Karkus2018, gu2015neuraladaptive}. NOPF training, therefore, does not require running a particle filter, differentiating through resampling, or using weighted particle trajectories as training targets. Once trained, the proposal is used as a plug-in replacement for the proposal step in a standard PF. During inference, each particle $\mathbf{x}_{t-1}^{(i)}$ and the new observation $\mathbf{y}_t$ are passed through the proposal network to produce Gaussian proposal parameters, from which $\mathbf{x}_t^{(i)}$ is sampled. The resulting particles are then weighted using Eq.~\ref{eqn:importance_sampling}, normalized, and resampled as usual. The training and inference procedures are summarized in Alg.~\ref{alg:training} and Alg.~\ref{alg:inference}, with full algorithmic details provided in Appendix~\ref{adx:training_and_inference}.

\begin{figure}[!htbp]
\centering

\begin{minipage}[t]{0.48\linewidth}
\begin{algorithm}[H]
\caption{Training}
\label{alg:training}
\footnotesize
\begin{algorithmic}[1]
\For{each iteration}
    \State simulate a batch of trajectories $\{(\mathbf{x}_{0:T}^{(b)},\mathbf{y}_{0:T}^{(b)})\}_{b=1}^B$
    \For{each $(\mathbf{x}_{t-1}, \mathbf{y}_t, t)$ in the batch}
        \State $(\boldsymbol{\mu}, \boldsymbol{\Sigma}) \gets q_\theta(\mathbf{x}_{t-1}, \mathbf{y}_t, t)$
        \State draw $S$ samples from $\mathbf{x}_{t}\sim\mathcal{N}(\boldsymbol{\mu}, \boldsymbol{\Sigma})$ and 
        \Statex\hspace{\algorithmicindent}\hspace{\algorithmicindent}use them to estimate $\mathcal{L}_{\text{ELBO}}(\theta; \mathbf{x}_{t-1}, \mathbf{y}_t, t)$
    \EndFor
    \State estimate $\mathcal{L}(\theta)$ over the batch
    \State $\theta \gets \theta - \eta\nabla_\theta \mathcal{L}({\theta})$
\EndFor
\end{algorithmic}
\end{algorithm}
\end{minipage}
\hfill
  \begin{minipage}[t]{0.48\linewidth}
  \begin{algorithm}[H]
  \caption{Inference}
  \label{alg:inference}
  \footnotesize
  \begin{algorithmic}[1]
  \State initialize $\mathbf{x}_0^{(i)}\sim p_0$, $w_0^{(i)}=1/N$
  \For{$t=1,\ldots,T$}
      \State $(\boldsymbol{\mu}_t^{(i)}, \boldsymbol{\Sigma}_t^{(i)}) \gets q_\theta(\mathbf{x}_{t-1}^{(i)}, \mathbf{y}_t, t)$
      \State sample $\mathbf{x}_t^{(i)} \sim \mathcal{N}(\boldsymbol{\mu}_t^{(i)}, \boldsymbol{\Sigma}_t^{(i)})$
      \State update weights using Eq.~\ref{eqn:importance_sampling}
      \State normalize weights to get $\sum_{i = 1}^N w_t^{(i)} = 1$
      \State systematic resample and reset $w_t^{(i)}=1/N$
      \vspace{1mm}
  \EndFor
  \vspace{1mm}
  \end{algorithmic}
  \end{algorithm}
  \end{minipage}

\end{figure}

\subsection{Evaluation Metrics}
\label{sec:metrics}

Since only a single realized trajectory $(\mathbf{x}_t^\star,\mathbf{y}_t)$ from the true system is available during rollout, evaluating the full filtering distribution is challenging. We therefore use several complementary metrics to evaluate different aspects of the particle approximation.

To assess state-estimation accuracy, we report normalized mean squared error (NMSE), which is most informative when the filtering distribution is approximately unimodal. To evaluate the quality of the full posterior approximation, particularly in multimodal settings where point estimates can be misleading, we use the one-step-ahead predictive negative log-likelihood (PNLL), derived from $p(\mathbf{y}_{t+1} \mid \mathbf{y}_{1:t})$, and the energy score (ES)~\cite{Gneiting2007}. PNLL measures how well the inferred posterior predicts future observations, while the energy score provides a bandwidth-free distributional score for comparing the full particle approximation:
\begin{align}
    \mathrm{ES}_t = \mathbb E_{\mathbf x_t\sim \widehat p_t}\|\mathbf x_t-\mathbf x_t^\star\|_2 - \frac12 \mathbb E_{\mathbf x_t,\mathbf x_t'\sim \widehat p_t}\|\mathbf x_t-\mathbf x_t'\|_2.
\label{eqn:energy_score}
\end{align}
Here $\widehat p_t$ denotes the empirical filtering distribution represented by the ensemble. The first term favors concentration near the realized state, while the second penalizes excessive ensemble spread.

To assess proposal efficiency, we report the pre-resampling effective sample size (ESS), $N_t^{\mathrm{eff}} = 1/\sum_{i=1}^N (w_t^{(i)})^2$, which quantifies particle-weight degeneracy~\cite{Arulampalam2002}. Finally, we report wall time (WT) in seconds for the full run to quantify computational cost.

For each example, we therefore report five time-averaged metrics: NMSE for estimation accuracy, PNLL and ES for distributional quality, ESS for proposal efficiency, and WT for computational cost. All computations, including forward passes through the learned proposal network, are performed on a CPU to maintain consistency across particle filters. Details of metric computation, additional diagnostics, and the computational setup are provided in Appendices~\ref{adx:metrics_computation} and~\ref{adx:comp_arch}.

\section{Experiments}
\label{sec:experiments}
We evaluate NOPF on robotics localization benchmarks of increasing dimensionality and inference complexity, from approximately unimodal settings to ambiguous multimodal problems. These benchmarks assess estimation accuracy, proposal efficiency, robustness to posterior multimodality, and computational cost. All filters are evaluated on trajectories simulated from the true stochastic model, with results reported as mean $\pm$ standard deviation across 10 random seeds. Angular state components use wrapped arithmetic throughout.

We compare against Kalman-based baselines when applicable, including KF for the linear-Gaussian benchmark and EnKF for nonlinear benchmarks. For the linear-Gaussian benchmark, we also include an Optimal Particle Filter (OPF), which samples from the closed-form locally optimal proposal. We further compare against standard and proposal-enhanced PF baselines, including the Bootstrap Particle Filter (BPF), Auxiliary Particle Filter (APF), Unscented Particle Filter (UPF), Nudged Particle Filter (NuPF), and Inner-Sampling Proposal Filter (ISPF). ISPF is an expensive local proposal baseline that uses an inner Monte Carlo approximation of the locally optimal proposal to sample the next particle for each ancestor. Implementation details for all filters are provided in Appendix~\ref{adx:filter_details}. NOPF is evaluated using the metrics in Section~\ref{sec:metrics}, with additional results, particle-count sensitivity studies, and supplementary metrics provided in Appendix~\ref{adx:additional_results} and~\ref{adx:metrics_computation}.

\subsection{Linear System with Biased Sensors}
\label{sec:ex1}
The first benchmark is a linear-Gaussian sensor-bias model for a robot moving along one dimension, used as a controlled setting in which the optimal Bayesian update is analytically tractable, and the Kalman filter provides an exact reference. The 3-dimensional latent state is $\mathbf{x}_t=[p_t,v_t,b_t]^\top$, where $p_t$ is position, $v_t$ velocity, and $b_t$ an additive bias in the velocity observations. The dynamics and observation model are
\begin{align}
\mathbf{x}_t =
\begin{bmatrix}
1 & \Delta t & 0 \\
0 & 1 & 0 \\
0 & 0 & 1
\end{bmatrix}
\mathbf{x}_{t-1}
+
\boldsymbol{\epsilon}_t,
\qquad
\mathbf{y}_t =
\begin{bmatrix}
1 & 0 & 0 \\
0 & 1 & 1
\end{bmatrix}
\mathbf{x}_t
+
\boldsymbol{\eta}_t,
\end{align}
with $\Delta t=0.1$, $\boldsymbol{\epsilon}_t\sim\mathcal N(0,\sigma^2 \mathbf I_3)$, and $\boldsymbol{\eta}_t\sim\mathcal N(0,\sigma^2 \mathbf I_2)$, where $\sigma=0.1$. The system is simulated from $t=1$ to $t=60$ with $\mathbf{x}_0=[1,0,0]^\top$.

\begin{figure}[!htbp]
    \centering
    \includegraphics[width=1\linewidth]{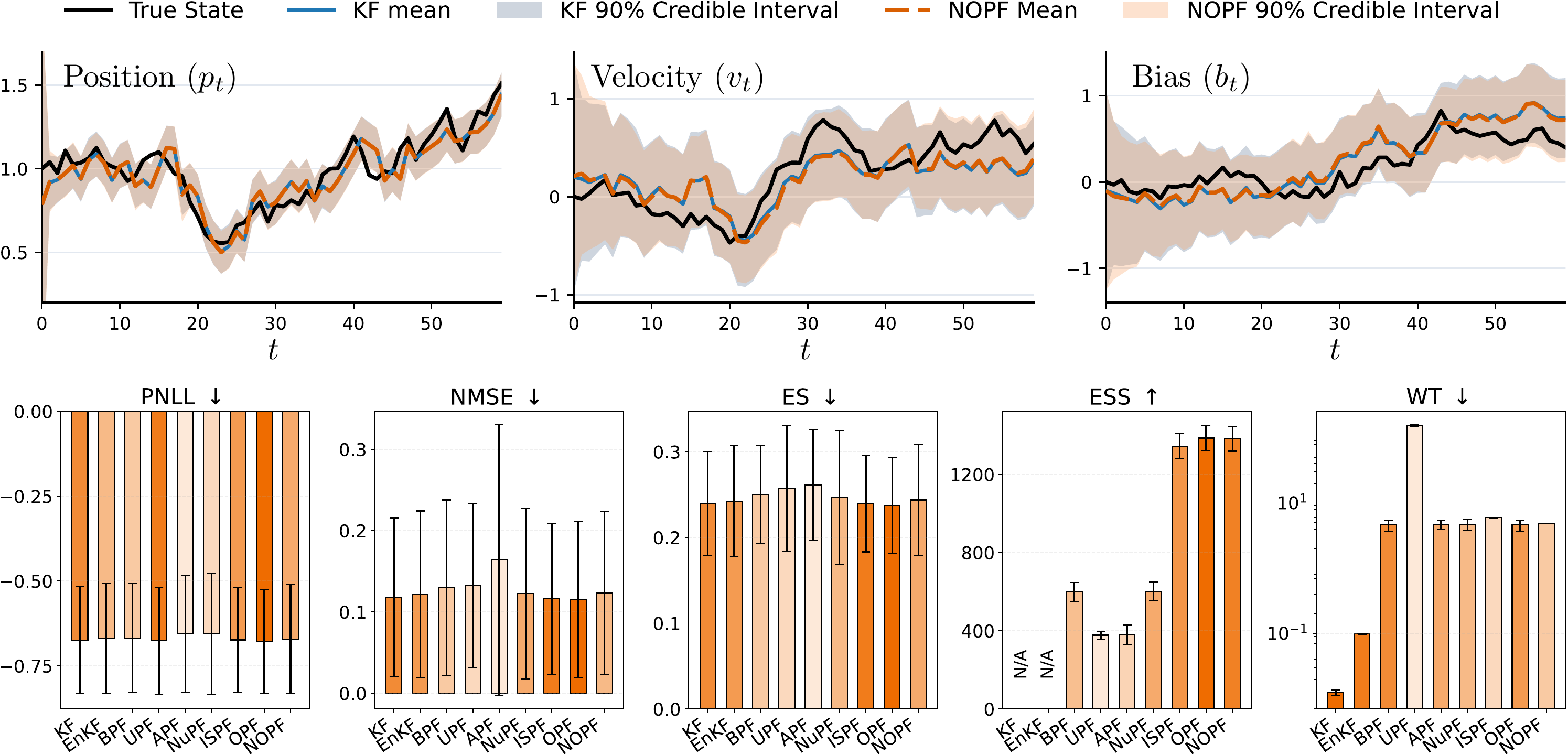}
    \caption{Linear system with biased sensors. Top: true and estimated trajectories with 90\% credible intervals for a representative seed. Bottom: aggregate metrics across 10 seeds. Arrows indicate whether larger or smaller values are preferred, and darker shading denotes better performance.}
    \label{fig:example_1}
\end{figure}

State trajectories for a representative seed and aggregate results over 10 seeds are shown in Fig.~\ref{fig:example_1}. All PFs and EnKF use $N=2000$ particles. As expected in this linear-Gaussian setting, all methods achieve NMSE and PNLL values comparable to those of the Kalman Filter. OPF samples from the closed-form locally optimal proposal and therefore provide a direct reference for optimal-proposal behavior. NOPF closely matches OPF and ISPF in estimation accuracy and ESS, while maintaining wall time comparable to simple PF baselines. The strong overlap of the 90\% credible intervals and agreement across metrics indicate that NOPF closely approximates optimal-proposal behavior in this setting, substantially improving sample efficiency with minimal computational overhead.

\subsection{2D Robot Localization with Range and Bearing Observations}
\label{sec:ex2}
The second benchmark is a 2D nonlinear localization problem with four known landmarks. The 7-dimensional state is $\mathbf x_t=[p_t^x,p_t^y,\theta_t,v_t,\omega_t,b_t^v,b_t^\omega]^\top$, where $p_t^x$ and $p_t^y$ denote robot position, $\theta_t$ heading, $v_t$ and $\omega_t$ linear and angular velocities, and $b_t^v,b_t^\omega$ odometry biases. The transition dynamics follow a nonlinear unicycle-type motion with a first-order velocity response to time-varying controls. The observations are represented as a 10-dimensional vector comprising noisy range and bearing measurements for all four landmarks, along with biased odometry data. Full model details are provided in Appendix~\ref{adx:benchmark_2}.

\begin{figure}[!htbp]
    \centering
    \includegraphics[width=1\linewidth]{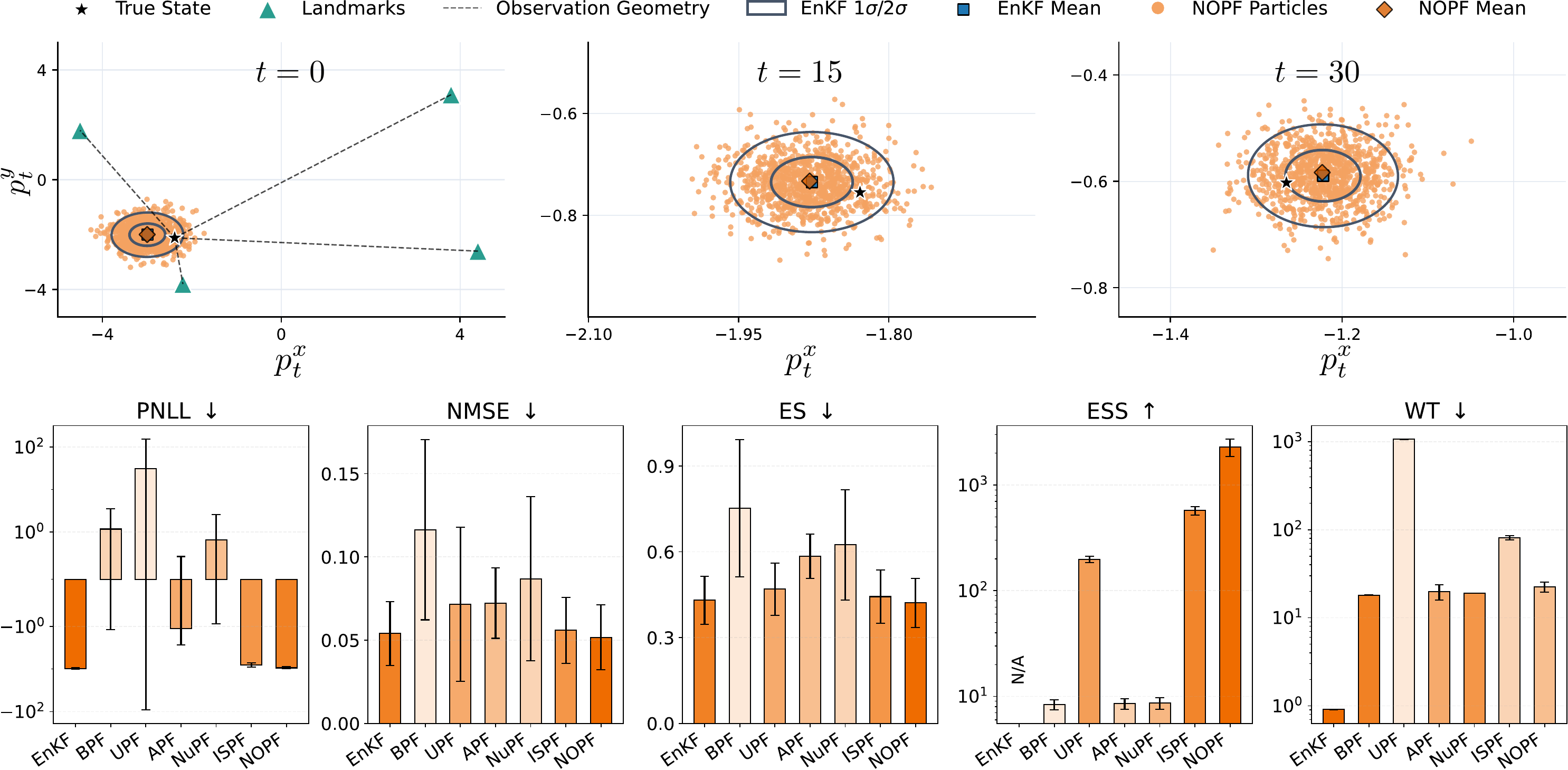}
    \caption{2D robot localization with range and bearing observations. Top: initial state and later-time posterior summaries for a representative seed with NOPF particles and the EnKF covariance approximation. Bottom: aggregate metrics across 10 seeds.}
    \label{fig:example_2}
\end{figure}

Representative snapshots at selected times from a single trajectory, together with aggregate results over 10 seeds, are shown in Fig.~\ref{fig:example_2}. All filters use $N=10,000$ particles. Because the range and bearing observations produce approximately unimodal filtering distributions, EnKF performs particularly well and provides a strong Gaussian baseline. NOPF achieves NMSE, PNLL, and ES comparable to EnKF while yielding substantially higher ESS than the other PF baselines. ISPF also improves over standard PF proposals but is considerably more expensive due to its inner Monte Carlo sampling. These results show that NOPF improves proposal efficiency in a nonlinear but approximately unimodal setting, without the computational cost of inner sampling or sigma-point-based proposals.

\subsection{2D Robot Localization with Biased Range-Only Observations}
\label{sec:ex3}
This benchmark employs dynamics similar to the previous example, but observations are now limited to a 3-dimensional vector comprising a noisy, biased range measurement from a landmark at the origin, together with noisy, biased odometry measurements for linear and angular velocity. The range bias $b_t^r$ follows a random walk and is incorporated into the state, yielding the 8-dimensional state $\mathbf x_t = [p_t^x, p_t^y, \theta_t, v_t,\omega_t, b_t^r, b_t^v, b_t^\omega]^\top$. This reduced observation information induces the multimodal ambiguity absent in the previous example. Full forward and observation model details are provided in Appendix~\ref{adx:benchmark_3}.

Because a single range measurement constrains position to a circle, the initial filtering prior is specified as a mirrored two-component mixture prior along the $x$-coordinate. As the system evolves, the transition dynamics eventually make one branch more consistent with the observation sequence. Therefore, this benchmark tests whether a filter can preserve and subsequently resolve a discrete multimodal ambiguity rather than prematurely collapsing onto an incorrect mode.

\begin{figure}[!htbp]
    \centering
    \includegraphics[width=1\linewidth]{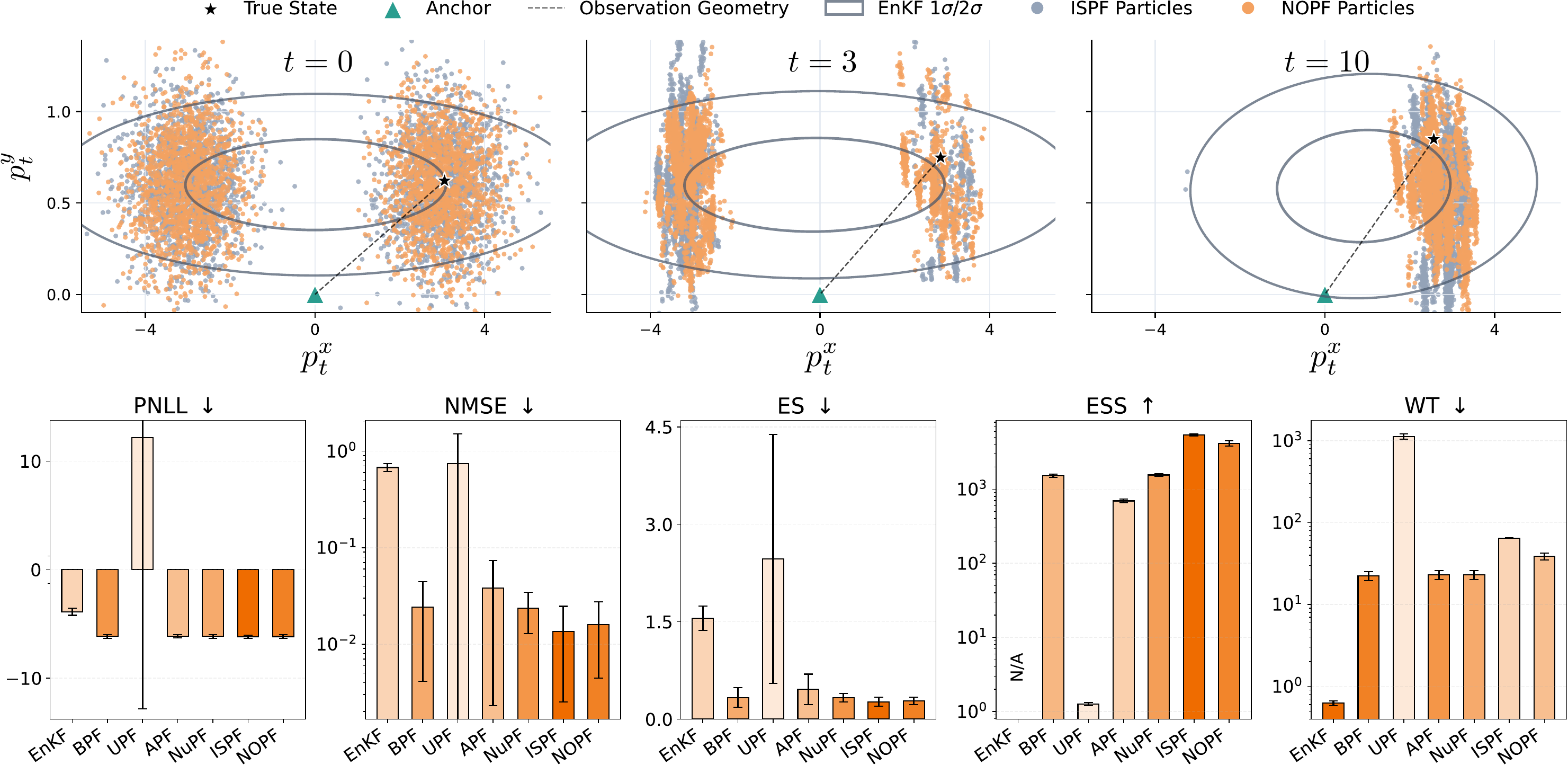}
    \caption{2D robot localization with biased range-only observations. Top: posterior snapshots showing whether filters preserve and resolve posterior branch ambiguity over time for a representative seed. Bottom: aggregate metrics across 10 seeds.}
    \label{fig:example_3}
\end{figure}

Representative snapshots at selected times from a single trajectory, together with aggregate results over 10 seeds using $N=10{,}000$ particles, are shown in Fig.~\ref{fig:example_3}. This example directly probes multimodal posterior tracking. The EnKF Gaussian approximation spreads probability mass across separated modes, producing an overly diffuse approximation that does not faithfully represent the bimodal posterior. NOPF preserves and resolves the multimodal structure more reliably than the competing filters, improving distributional accuracy as measured by PNLL and ES while also achieving high ESS. ISPF provides a strong inner-sampling approximation to the locally optimal proposal, but is substantially more expensive. These results indicate that the learned proposal improves both sample efficiency and multimodal posterior tracking when observations are weak and ambiguous.

\subsection{3D Robot Localization with Unordered Range-Only Observations}
\label{sec:ex4}
The final benchmark is a 20-dimensional nonlinear 3D localization problem with four fixed anchors lying on the plane $z=0$. Unlike the previous range-only benchmark, where observations consist of labeled ranges, the filter now observes an unordered set of four range returns and does not know which anchor generated which measurement. This introduces data-association ambiguity in addition to state uncertainty. The observation also includes three proprioceptive channels, consisting of biased speed, yaw-rate, and pitch-rate measurements, yielding a 7-dimensional observation vector. The latent state includes 3D position, orientation, motion states, proprioceptive and range biases, slowly varying drift terms representative of wind disturbances, and multiplicative calibration errors. Full model details are provided in Appendix~\ref{adx:benchmark_4}. In the PF likelihood, anchor identity is handled by marginalizing over all $4!$ measurement-to-anchor assignments.

The filtering prior is specified as a two-component Gaussian mixture. Together with the unlabeled measurements, this results in a challenging, high-dimensional, multimodal posterior with multiple plausible branches. This benchmark, therefore, tests whether a filter can preserve and eventually resolve multimodal ambiguities rather than collapse prematurely in high-dimensional settings.

\begin{figure}[!htbp]
    \centering
    \includegraphics[width=\linewidth]{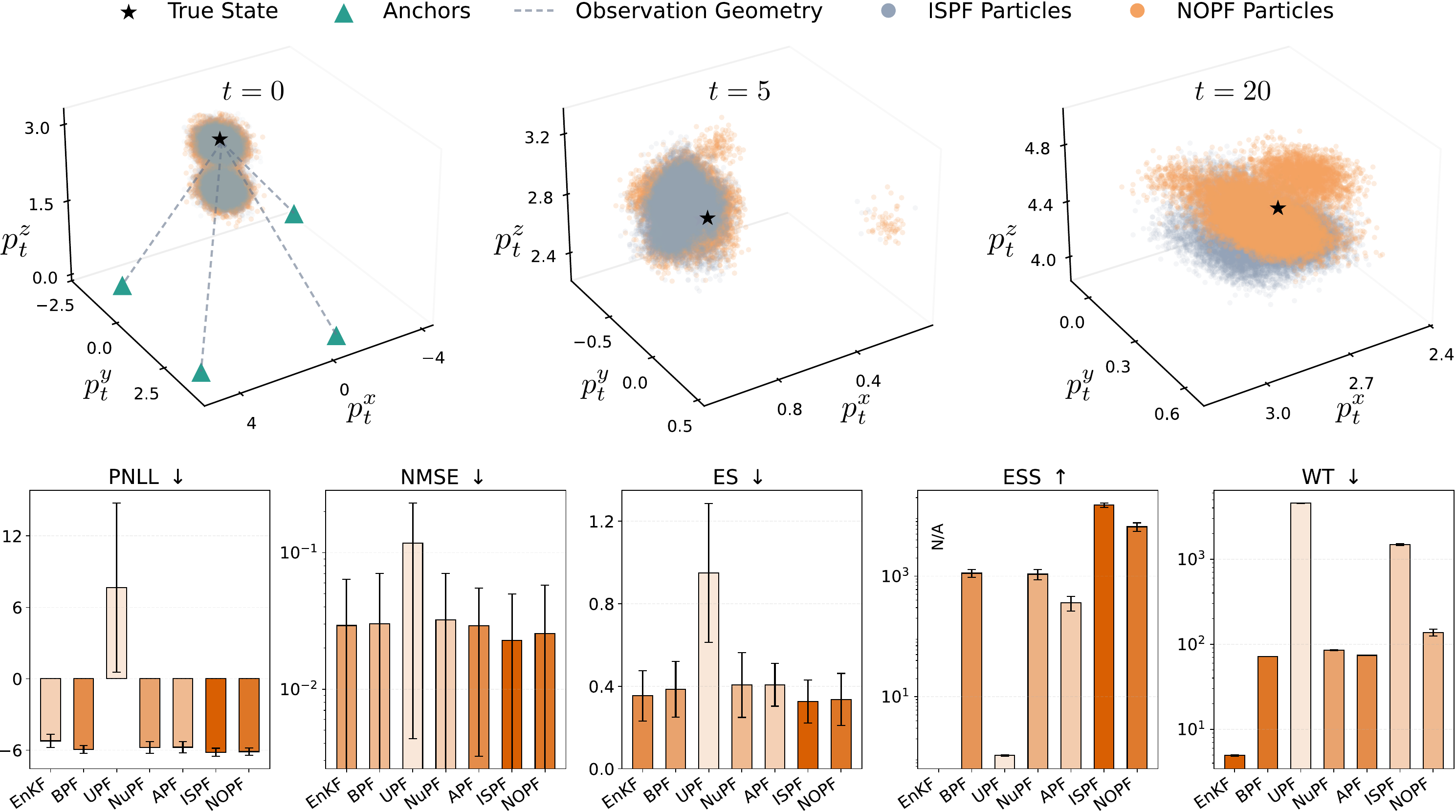}
    \caption{3D robot localization with unordered range-only observations. Top: snapshots showing the evolution of multimodal filtering distribution for a representative seed. Bottom: aggregate metrics across 10 seeds.}
    \label{fig:example_4}
\end{figure}

Representative snapshots at selected times from a single trajectory, together with aggregate results over 10 seeds using $N=50,000$ particles, are shown in Fig.~\ref{fig:example_4}. This is the most challenging benchmark considered, jointly stressing proposal quality, multimodal posterior tracking, and robustness to latent data association. NOPF improves PNLL and ES over standard PF baselines while maintaining high ESS and moderate computational cost. ISPF achieves comparable distributional accuracy and higher ESS, suggesting that the local proposal structure is more complex than a single Gaussian in this benchmark. Nevertheless, NOPF achieves similar performance with substantially lower wall time, highlighting the robustness of the Gaussian-learned proposal as a computationally efficient approximation.

\section{Discussion and Limitations}
\label{sec:discussion}
We introduced Neural Optimal Particle Filters (NOPFs), which combine classical PFs with learned proposal mechanisms while preserving the sequential importance sampling framework. NOPF learns an amortized approximation to the locally optimal proposal. This provides a principled local target because, within recursive one-step PF updates where the current observation is used to sample the new particle but does not alter the previous particle locations, the locally optimal proposal gives the least-degenerate benchmark~\cite{Snyder2015}. This remains true even when richer proposal classes are allowed to condition on particle histories or the previous ensemble~\cite{Doucet2000, vanLeeuwen2019}. By biasing particles toward states that are both dynamically plausible and observation-consistent, NOPF reduces weight degeneracy with modest additional cost, primarily from a neural network forward pass and Gaussian density evaluation. Because the learned proposal is used in the same place as any PF proposal and is still corrected by the standard importance weights, NOPF improves particle placement without changing the target filtering distribution.

Across the benchmarks, NOPF achieves a favorable tradeoff between estimation accuracy, sample efficiency, and computational cost. It matches Kalman-based and Gaussian baselines in approximately unimodal settings and outperforms standard PF baselines in multimodal problems. The gains from learned proposals become more pronounced as observational ambiguity and posterior multimodality increase. Since the proposal is amortized offline, inference requires no online optimization or sigma-point construction beyond a neural network forward pass, avoiding computational bottlenecks associated with methods such as UPF and ISPF.

The Gaussian conditional proposal used here does not prevent the overall filtering distribution $p(\mathbf{x}_t \mid \mathbf{y}_{1:t})$ from representing multimodal beliefs. The Gaussian assumption applies only to the proposal conditioned on a single ancestor particle $\mathbf{x}_{t-1}$. After combining these conditional proposals over the weighted particle ensemble, the resulting posterior approximation is effectively a mixture and can remain highly non-Gaussian. Conditioning on $\mathbf{x}_{t-1}$ often makes the local proposal substantially simpler than the full posterior, motivating the Gaussian parameterization as a useful inductive bias.

The method has three main limitations. First, the proposal network is trained offline using trajectories generated from the forward model. This can lead to train-test mismatch if the filtering distribution encountered during inference drifts outside the training distribution, or if the simulator does not faithfully represent the true system. In such cases, proposal quality may degrade. However, as long as the proposal has adequate support, the standard importance-weight update preserves the target distribution in the usual importance-sampling sense, and poor proposals primarily appear through increased weight variance rather than asymptotic bias.

Second, the current training implementation uses reparameterized samples from the learned proposal to optimize the ELBO. This requires differentiable evaluations of the transition density and likelihood with respect to the proposed state. Systems with non-differentiable simulators or observation operators would require alternative gradient estimators or differentiable surrogate likelihood models.

Third, the Gaussian conditional proposal may be restrictive when the locally optimal proposal for a single ancestor particle is strongly multimodal, as under severe data-association ambiguity. In such cases, a unimodal proposal may average across modes and place samples in low-probability regions, thereby increasing the variance of the importance weights. The results suggest that the Gaussian approximation is robust in the benchmarks considered, but richer tractable proposal families, such as mixtures or normalizing flows, may be beneficial in more strongly multimodal local proposal settings.

Overall, these results suggest that learned proposal mechanisms can improve the practicality of PFs without sacrificing principled Bayesian weighting. Future work includes more expressive proposal families and online adaptation mechanisms to reduce the train-test distribution shift.

\newpage



{
\small
\bibliographystyle{unsrt}
\bibliography{ref}
}

%



\appendix

\section{Additional Results}
\label{adx:additional_results}
This section reports supplementary results supporting the main text. We examine sensitivity to the number of particles and include additional diagnostics not shown in the main paper.

\subsection{Metrics vs Number of Particles}
\label{adx:metrics_vs_n}

\begin{figure}[!htbp]
    \centering
    \includegraphics[width=\linewidth]{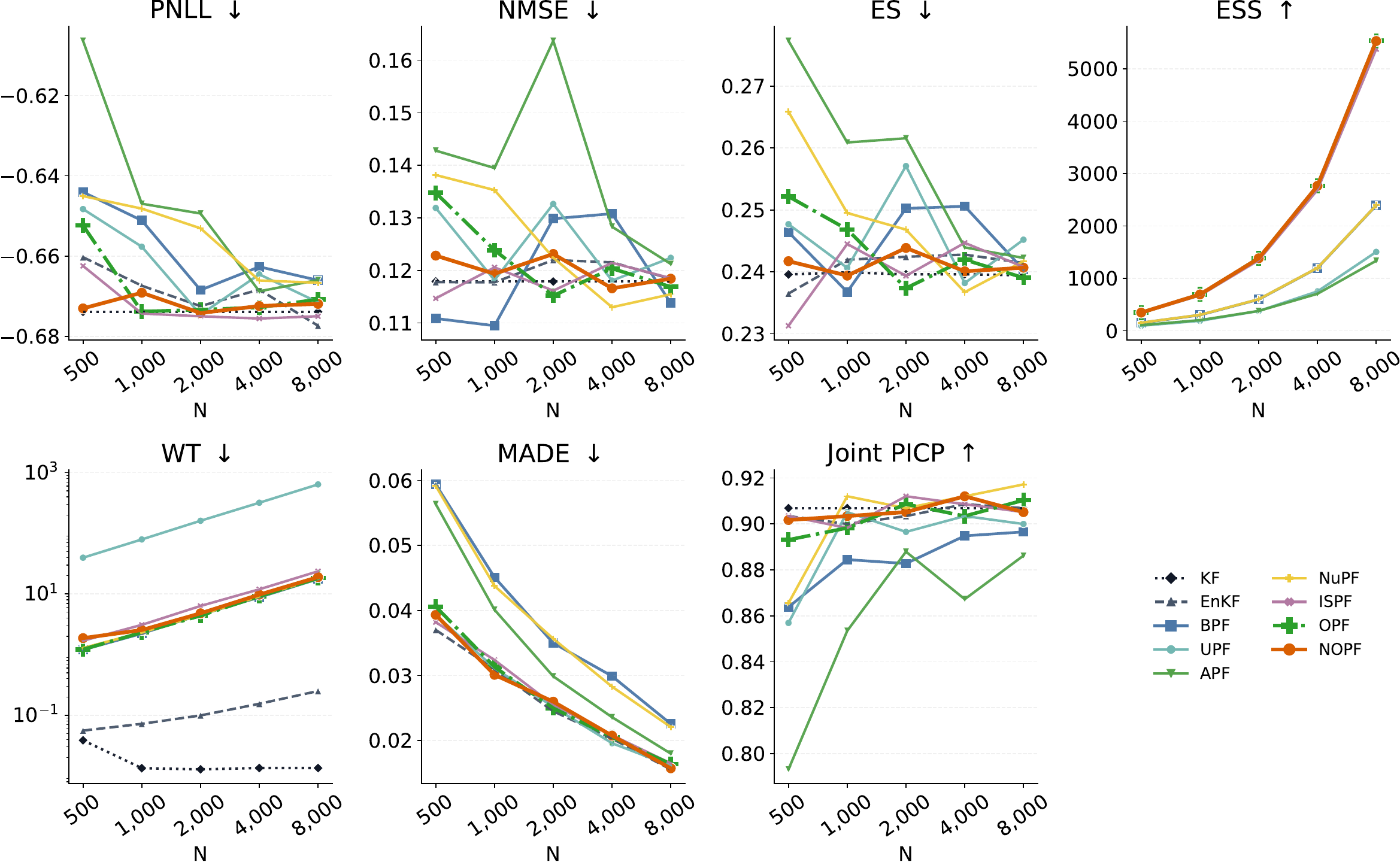}
    \caption{Linear system with biased sensors: Metrics as a function of the number of particles $N$.}
    \label{fig:ex1_vs_n}
\end{figure}

\begin{figure}[!htbp]
    \centering
    \includegraphics[width=\linewidth]{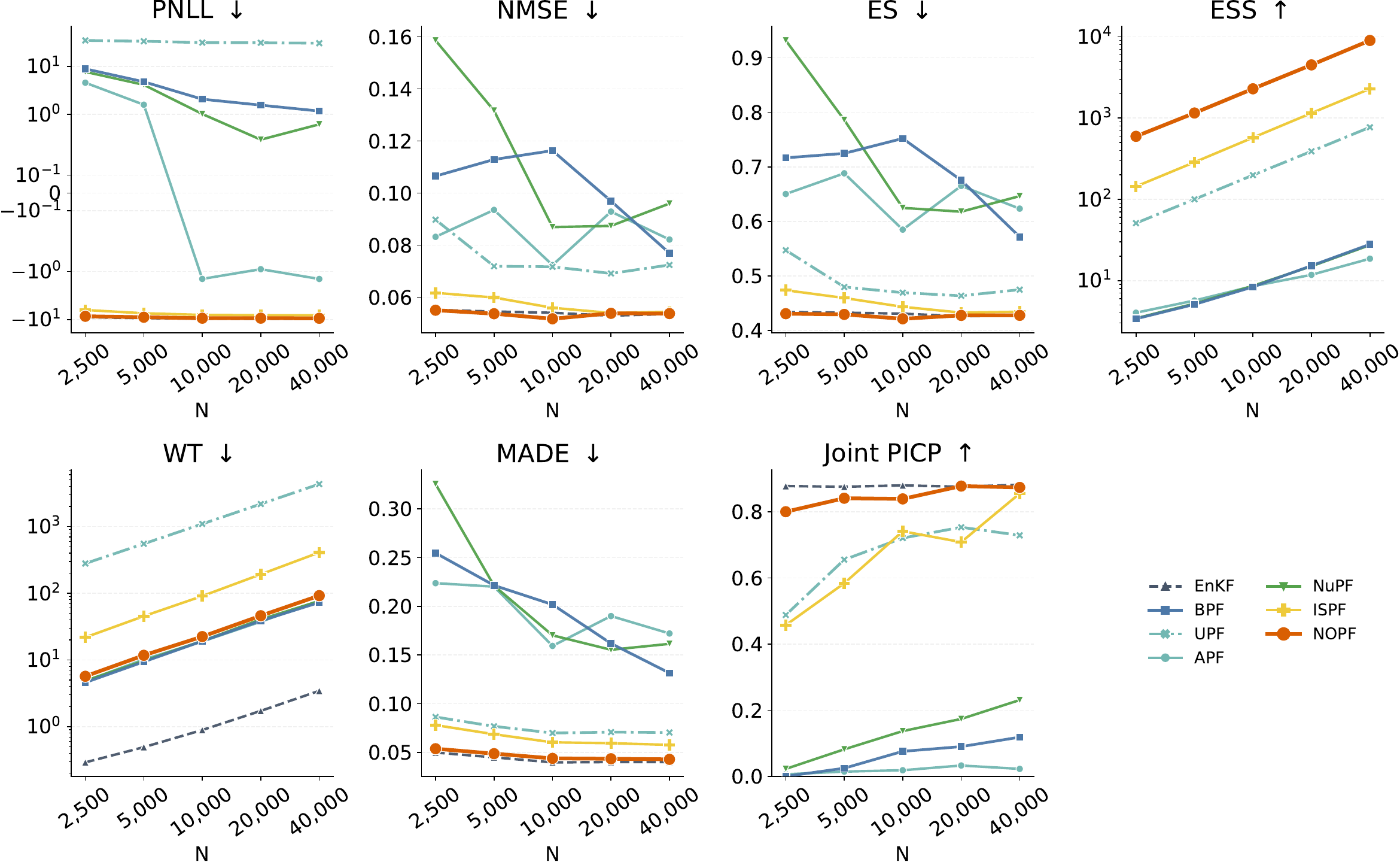}
    \caption{2D robot localization with range and bearing observations: Metrics as a function of the number of particles $N$.}
    \label{fig:ex2_vs_n}
\end{figure}

\begin{figure}[!htbp]
    \centering
    \includegraphics[width=\linewidth]{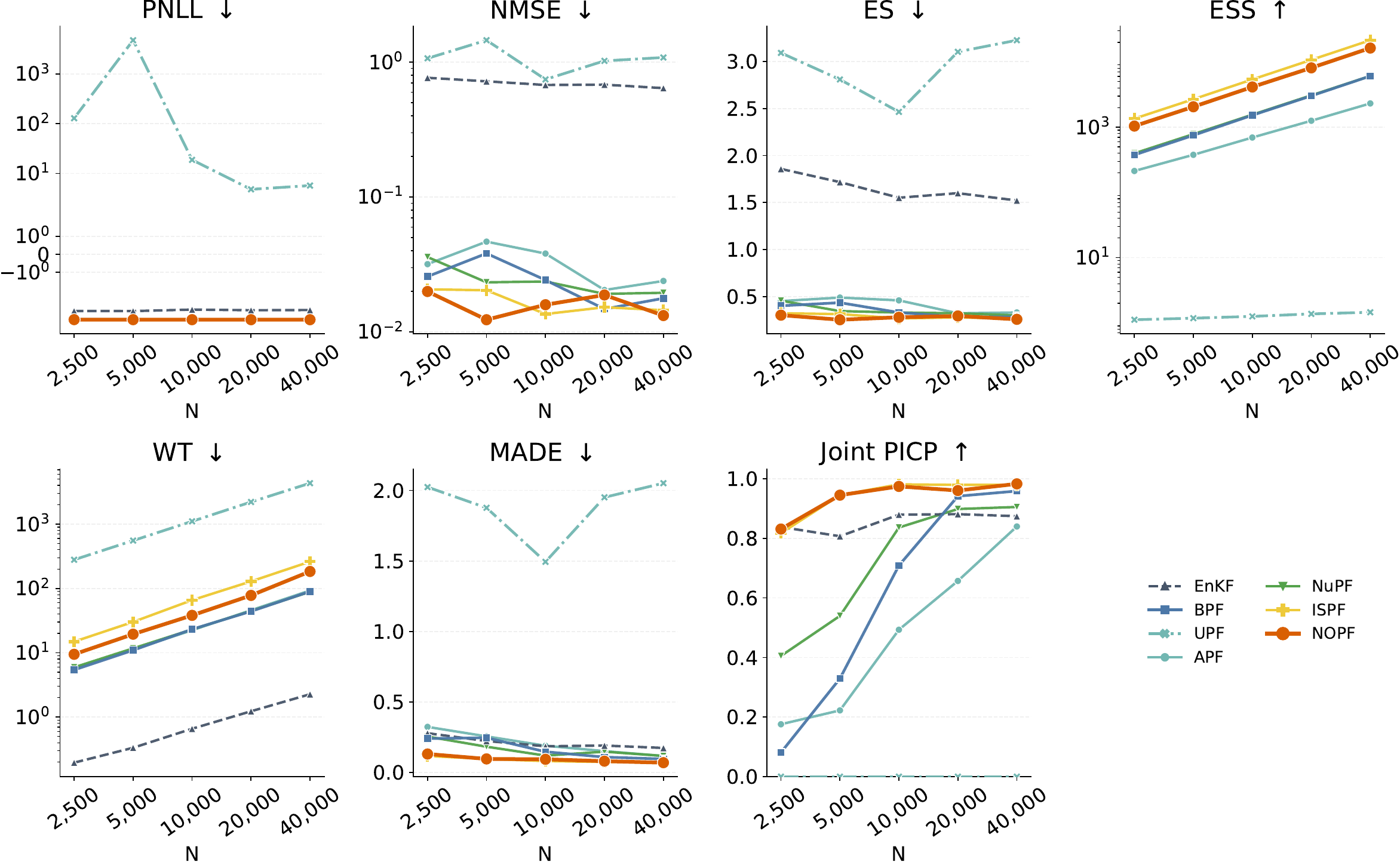}
    \caption{2D robot localization with biased range-only observations: Metrics as a function of the number of particles $N$.}
    \label{fig:ex3_vs_n}
\end{figure}

\begin{figure}[!htbp]
    \centering
    \includegraphics[width=\linewidth]{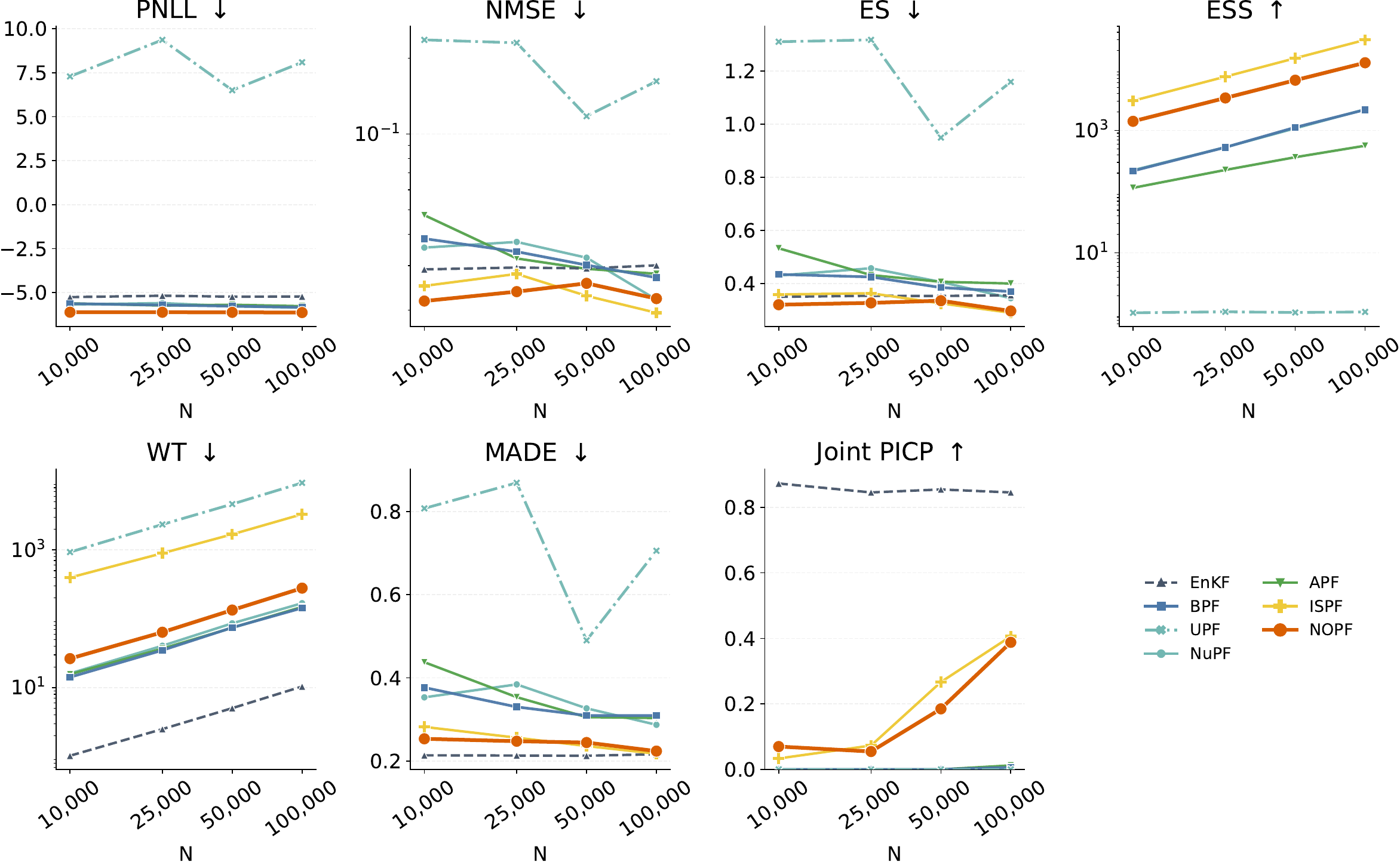}
    \caption{3D robot localization with unordered range-only observations: Metrics as a function of the number of particles $N$.}
    \label{fig:ex4_vs_n}
\end{figure}

We report the variation of the metrics introduced in Section~\ref{sec:metrics} and Appendix~\ref{adx:additional_metrics} as the number of particles $N$ is varied, for all benchmarks in Fig.~\ref{fig:ex1_vs_n}--\ref{fig:ex4_vs_n}. All results are averaged over 10 random seeds. Across all benchmarks, NOPF maintains consistently higher effective sample sizes and competitive or improved accuracy across a wide range of particle counts.

\subsection{Tabulated Results}
\label{adx:tab_results_1}
This section reports complete per-filter results for each of the four benchmarks introduced in Section~\ref{sec:experiments}. For each benchmark, we evaluate the applicable filters described in Appendix~\ref{adx:filter_details}. For the linear-Gaussian sensor-bias benchmark, we additionally include the Kalman Filter and the Optimal Particle Filter. Each scalar metric is computed per seed using the procedure in Appendix~\ref{adx:metrics_computation} and aggregated across $10$ independent random seeds; we report the across-seed mean $\pm$ sample standard deviation in the format $\mu\pm\sigma$. 

The metric set in each table extends the five summary metrics highlighted in the main text with the two distributional diagnostics from Appendix~\ref{adx:additional_metrics}, giving seven metrics per filter: \textbf{NMSE} ($\downarrow$), \textbf{PNLL} ($\downarrow$), \textbf{ES} ($\downarrow$), \textbf{ESS} ($\uparrow$), \textbf{WT} ($\downarrow$), \textbf{MADE} ($\downarrow$), and \textbf{Joint PICP} ($\to 0.9$). The arrow next to each header in Tables~\ref{tab:adx-results-sbs}--\ref{tab:adx-results-unordered-range} indicates the preferred direction. For Joint PICP, the target value is $0.9$.

\begin{table}[H]
\centering
\small
\setlength{\tabcolsep}{4pt}
\renewcommand{\arraystretch}{1.15}
\caption{Mean $\pm$ standard deviation across 10 seeds for the linear system with biased sensors benchmark (Section~\ref{sec:ex1}). Bold marks the top two per column (closest to $0.9$ for Joint PICP).}
\label{tab:adx-results-sbs}
\resizebox{\textwidth}{!}{%
\begin{tabular}{lccccccc}
\toprule
\textbf{Filter} & \textbf{NMSE} $\downarrow$ & \textbf{PNLL} $\downarrow$ & \textbf{ES} $\downarrow$ & \textbf{ESS} $\uparrow$ & \textbf{WT [s]} $\downarrow$ & \textbf{MADE} $\downarrow$ & \textbf{Joint PICP} $\to 0.9$ \\
\midrule
KF   & $0.1179 \pm 0.0973$ & $-0.6739 \pm 0.1572$ & $0.2396 \pm 0.0601$ & -- & $\mathbf{0.0124 \pm 0.0012}$ & -- & $0.9069 \pm 0.0502$ \\
EnKF & $0.1220 \pm 0.1022$ & $-0.6696 \pm 0.1616$ & $0.2424 \pm 0.0645$ & -- & $\mathbf{0.0990 \pm 0.0018}$ & $\mathbf{0.0245 \pm 0.0033}$ & $\mathbf{0.9034 \pm 0.0558}$ \\
BPF  & $0.1298 \pm 0.1079$ & $-0.6682 \pm 0.1608$ & $0.2502 \pm 0.0572$ & $598.5094 \pm 47.3294$ & $4.5919 \pm 0.9028$ & $0.0350 \pm 0.0054$ & $0.8828 \pm 0.0634$ \\
UPF  & $0.1326 \pm 0.1007$ & $\mathbf{-0.6764 \pm 0.1579}$ & $0.2571 \pm 0.0736$ & $377.0515 \pm 19.4465$ & $154.5473 \pm 3.1899$ & $0.0252 \pm 0.0029$ & $\mathbf{0.8966 \pm 0.0685}$ \\
APF  & $0.1637 \pm 0.1663$ & $-0.6559 \pm 0.1727$ & $0.2616 \pm 0.0647$ & $377.9505 \pm 50.1158$ & $4.6464 \pm 0.6950$ & $0.0299 \pm 0.0032$ & $0.8879 \pm 0.0547$ \\
NuPF & $0.1224 \pm 0.1051$ & $-0.6561 \pm 0.1791$ & $0.2468 \pm 0.0779$ & $601.1692 \pm 48.0596$ & $4.6895 \pm 0.9368$ & $0.0357 \pm 0.0076$ & $0.9069 \pm 0.0714$ \\
ISPF & $\mathbf{0.1161 \pm 0.0930}$ & $-0.6736 \pm 0.1556$ & $\mathbf{0.2394 \pm 0.0561}$ & $1345.2554 \pm 65.1092$ & $5.9684 \pm 0.0435$ & $0.0255 \pm 0.0027$ & $0.9121 \pm 0.0542$ \\
OPF  & $\mathbf{0.1152 \pm 0.0956}$ & $\mathbf{-0.6776 \pm 0.1531}$ & $\mathbf{0.2374 \pm 0.0558}$ & $\mathbf{1385.6603 \pm 64.2703}$ & $4.5959 \pm 0.9104$ & $\mathbf{0.0250 \pm 0.0031}$ & $0.9086 \pm 0.0539$ \\
NOPF & $0.1231 \pm 0.1001$ & $-0.6707 \pm 0.1594$ & $0.2439 \pm 0.0649$ & $\mathbf{1382.6653 \pm 64.0090}$ & $4.8077 \pm 0.0344$ & $0.0260 \pm 0.0037$ & $0.9052 \pm 0.0522$ \\
\bottomrule
\end{tabular}%
}
\end{table}

\begin{table}[H]
\centering
\small
\setlength{\tabcolsep}{4pt}
\renewcommand{\arraystretch}{1.15}
\caption{Mean $\pm$ standard deviation across 10 seeds for the 2D robot localization benchmark with range and bearing observations (Section~\ref{sec:ex2}). Bold marks the top two per column (closest to $0.9$ for Joint PICP).}
\label{tab:adx-results-rb}
\resizebox{\textwidth}{!}{%
\begin{tabular}{lccccccc}
\toprule
\textbf{Filter} & \textbf{NMSE} $\downarrow$ & \textbf{PNLL} $\downarrow$ & \textbf{ES} $\downarrow$ & \textbf{ESS} $\uparrow$ & \textbf{WT [s]} $\downarrow$ & \textbf{MADE} $\downarrow$ & \textbf{Joint PICP} $\to 0.9$ \\
\midrule
EnKF & $\mathbf{0.0540 \pm 0.0193}$ & $\mathbf{-9.6921 \pm 0.4545}$ & $\mathbf{0.4309 \pm 0.0838}$ & -- & $\mathbf{0.9094 \pm 0.0094}$ & $\mathbf{0.0397 \pm 0.0039}$ & $\mathbf{0.8796 \pm 0.0863}$ \\
BPF  & $0.1163 \pm 0.0541$ & $1.1781 \pm 2.3617$ & $0.7519 \pm 0.2399$ & $8.3675 \pm 0.9149$ & $\mathbf{18.0811 \pm 0.0524}$ & $0.2018 \pm 0.1127$ & $0.0755 \pm 0.0962$ \\
UPF  & $0.0717 \pm 0.0463$ & $31.3362 \pm 122.3115$ & $0.4694 \pm 0.0915$ & $198.3859 \pm 13.3273$ & $1064.7424 \pm 2.5994$ & $0.0698 \pm 0.0786$ & $0.7204 \pm 0.2411$ \\
APF  & $0.0723 \pm 0.0212$ & $-1.1114 \pm 1.5949$ & $0.5849 \pm 0.0773$ & $8.5483 \pm 0.9656$ & $19.8896 \pm 3.9272$ & $0.1593 \pm 0.0165$ & $0.0184 \pm 0.0326$ \\
NuPF & $0.0869 \pm 0.0493$ & $0.8330 \pm 1.7710$ & $0.6247 \pm 0.1931$ & $8.6296 \pm 1.0483$ & $19.0552 \pm 0.0677$ & $0.1702 \pm 0.0448$ & $0.1367 \pm 0.0545$ \\
ISPF & $0.0559 \pm 0.0198$ & $-8.0896 \pm 0.9456$ & $0.4434 \pm 0.0930$ & $\mathbf{574.4442 \pm 52.8567}$ & $81.0166 \pm 4.3176$ & $0.0603 \pm 0.0071$ & $0.7408 \pm 0.1902$ \\
NOPF & $\mathbf{0.0517 \pm 0.0196}$ & $\mathbf{-9.3536 \pm 0.5214}$ & $\mathbf{0.4215 \pm 0.0855}$ & $\mathbf{2289.7282 \pm 429.9442}$ & $22.5374 \pm 3.0092$ & $\mathbf{0.0438 \pm 0.0055}$ & $\mathbf{0.8388 \pm 0.1804}$ \\
\bottomrule
\end{tabular}%
}
\end{table}

\begin{table}[H]
\centering
\small
\setlength{\tabcolsep}{4pt}
\renewcommand{\arraystretch}{1.15}
\caption{Mean $\pm$ standard deviation across 10 seeds for the 2D robot localization benchmark with biased range-only observations (Section~\ref{sec:ex3}). Bold marks the top two per column (closest to $0.9$ for Joint PICP).}
\label{tab:adx-results-rangeonly}
\resizebox{\textwidth}{!}{%
\begin{tabular}{lccccccc}
\toprule
\textbf{Filter} & \textbf{NMSE} $\downarrow$ & \textbf{PNLL} $\downarrow$ & \textbf{ES} $\downarrow$ & \textbf{ESS} $\uparrow$ & \textbf{WT [s]} $\downarrow$ & \textbf{MADE} $\downarrow$ & \textbf{Joint PICP} $\to 0.9$ \\
\midrule
EnKF & $0.6767 \pm 0.0642$ & $-3.0860 \pm 0.2682$ & $1.5510 \pm 0.1880$ & -- & $\mathbf{0.6236 \pm 0.0395}$ & $0.1856 \pm 0.0344$ & $\mathbf{0.8793 \pm 0.1044}$ \\
BPF  & $0.0242 \pm 0.0201$ & $-4.9084 \pm 0.1347$ & $0.3296 \pm 0.1510$ & $1525.8024 \pm 72.6469$ & $\mathbf{22.3428 \pm 2.6978}$ & $0.1467 \pm 0.0748$ & $0.7086 \pm 0.3181$ \\
UPF  & $0.7444 \pm 0.7660$ & $18.7829 \pm 41.4584$ & $2.4642 \pm 1.9179$ & $1.2591 \pm 0.0547$ & $1125.3255 \pm 84.5301$ & $1.4938 \pm 1.4605$ & $0.0000 \pm 0.0000$ \\
APF  & $0.0381 \pm 0.0358$ & $-4.8894 \pm 0.1309$ & $0.4589 \pm 0.2336$ & $694.8429 \pm 43.7893$ & $22.9959 \pm 2.8219$ & $0.1894 \pm 0.1233$ & $0.4931 \pm 0.2152$ \\
NuPF & $0.0236 \pm 0.0109$ & $-4.9030 \pm 0.1287$ & $0.3295 \pm 0.0631$ & $1556.0490 \pm 67.3410$ & $23.0519 \pm 3.0155$ & $0.1193 \pm 0.0307$ & $\mathbf{0.8362 \pm 0.2027}$ \\
ISPF & $\mathbf{0.0136 \pm 0.0111}$ & $\mathbf{-4.9203 \pm 0.1202}$ & $\mathbf{0.2664 \pm 0.0714}$ & $\mathbf{5399.6077 \pm 172.8422}$ & $64.7673 \pm 0.5461$ & $\mathbf{0.0805 \pm 0.0124}$ & $0.9810 \pm 0.0275$ \\
NOPF & $\mathbf{0.0159 \pm 0.0115}$ & $\mathbf{-4.9131 \pm 0.1221}$ & $\mathbf{0.2787 \pm 0.0550}$ & $\mathbf{4139.9681 \pm 345.6060}$ & $38.5837 \pm 3.6579$ & $\mathbf{0.0933 \pm 0.0230}$ & $0.9741 \pm 0.0219$ \\
\bottomrule
\end{tabular}%
}
\end{table}

\begin{table}[H]
\centering
\small
\setlength{\tabcolsep}{4pt}
\renewcommand{\arraystretch}{1.15}
\caption{Mean $\pm$ standard deviation across 10 seeds for the 3D robot localization benchmark with unordered range-only observations (Section~\ref{sec:ex4}). Bold marks the top two per column (closest to $0.9$ for Joint PICP).}
\label{tab:adx-results-unordered-range}
\resizebox{\textwidth}{!}{%
\begin{tabular}{lccccccc}
\toprule
\textbf{Filter} & \textbf{NMSE} $\downarrow$ & \textbf{PNLL} $\downarrow$ & \textbf{ES} $\downarrow$ & \textbf{ESS} $\uparrow$ & \textbf{WT [s]} $\downarrow$ & \textbf{MADE} $\downarrow$ & \textbf{Joint PICP} $\to 0.9$ \\
\midrule
EnKF & $0.0293 \pm 0.0348$ & $-5.2189 \pm 0.5531$ & $0.3537 \pm 0.1214$ & -- & $\mathbf{4.9304 \pm 0.0908}$ & $\mathbf{0.2125 \pm 0.0463}$ & $\mathbf{0.8545 \pm 0.3030}$ \\
BPF  & $0.0302 \pm 0.0401$ & $-5.9356 \pm 0.3396$ & $0.3854 \pm 0.1351$ & $1118.3518 \pm 170.3681$ & $\mathbf{71.7007 \pm 0.2993}$ & $0.3092 \pm 0.1128$ & $0.0000 \pm 0.0000$ \\
UPF  & $0.1176 \pm 0.1132$ & $7.6478 \pm 7.1106$ & $0.9492 \pm 0.3356$ & $1.0644 \pm 0.0272$ & $4545.2538 \pm 16.3267$ & $0.4902 \pm 0.2829$ & $0.0000 \pm 0.0000$ \\
NuPF & $0.0323 \pm 0.0381$ & $-5.7771 \pm 0.5003$ & $0.4065 \pm 0.1575$ & $1078.1617 \pm 213.1463$ & $85.0738 \pm 0.6555$ & $0.3266 \pm 0.1164$ & $0.0000 \pm 0.0000$ \\
APF  & $0.0291 \pm 0.0259$ & $-5.7518 \pm 0.4722$ & $0.4071 \pm 0.1036$ & $364.4146 \pm 100.0679$ & $74.0771 \pm 0.6063$ & $0.3053 \pm 0.1011$ & $0.0000 \pm 0.0000$ \\
ISPF & $\mathbf{0.0228 \pm 0.0270}$ & $\mathbf{-6.1685 \pm 0.3328}$ & $\mathbf{0.3259 \pm 0.1042}$ & $\mathbf{15050.6945 \pm 1188.3015}$ & $1476.2126 \pm 37.9838$ & $\mathbf{0.2359 \pm 0.0549}$ & $\mathbf{0.2667 \pm 0.2824}$ \\
NOPF & $\mathbf{0.0255 \pm 0.0325}$ & $\mathbf{-6.1240 \pm 0.3260}$ & $\mathbf{0.3360 \pm 0.1263}$ & $\mathbf{6584.2455 \pm 1056.6422}$ & $137.5812 \pm 12.2964$ & $0.2443 \pm 0.0946$ & $0.1848 \pm 0.2145$ \\
\bottomrule
\end{tabular}%
}
\end{table}

\section{Neural Proposal: Architecture and Training}
\label{adx:neural_proposal}

This appendix expands on the implementation of the learned proposal density introduced in Sec.~\ref{sec:method}. We give the explicit conditional-Gaussian parameterization used in all experiments (Appendix~\ref{adx:proposal_parameterization}), state the training objective (Appendix~\ref{adx:loss_function}), and describe the offline training and online filtering algorithms (Appendix~\ref{adx:training_and_inference}). All choices are common across benchmarks unless explicitly noted; benchmark-specific overrides (network sizes, clamping ranges, training hyperparameters) are summarized in Table~\ref{tab:adx-nn-hparams} and in Appendix~\ref{adx:experiment_details}. The metrics used to evaluate the resulting filter and the hardware on which they are reported are deferred to Appendices~\ref{adx:metrics_computation} and \ref{adx:comp_arch}, respectively.

\subsection{Parameterization of Mean and Covariance Matrices for the Learned Proposal}
\label{adx:proposal_parameterization}

The NeuralPF proposal is parameterized as the conditional Gaussian
\begin{align}
q_\theta(\mathbf{x}_t \mid \mathbf{x}_{t-1}, \mathbf{y}_t, t)
=
\mathcal{N}\!\left(
\mathbf{x}_t;
\boldsymbol{\mu}_\theta(\hat{\mathbf{x}}_t,\mathbf{y}_t,t),
L_{\theta,x}(\hat{\mathbf{x}}_t,\mathbf{y}_t,t)
L_{\theta,x}(\hat{\mathbf{x}}_t,\mathbf{y}_t,t)^\top
\right),
\label{eq:adx-neural-proposal}
\end{align}
where $\hat{\mathbf{x}}_t=f(\mathbf{x}_{t-1},t-1)$ is the deterministic noise-free propagation of the previous state. The network input is
\begin{align}
\mathbf{u}_t = [\hat{\mathbf{x}}_t,\mathbf{y}_t,t] \in \mathbb{R}^{d_x+d_y+1}.
\label{eq:adx-network-input}
\end{align}
Using $\hat{\mathbf{x}}_t$ avoids requiring the network to relearn the forward model and instead focuses learning on the observation-informed proposal correction.

The proposal is parameterized in whitened residual coordinates. Let $Q$ denote the forward-noise covariance and let $\boldsymbol{\sigma}_f=\sqrt{\operatorname{diag}(Q)}$ denote the vector of forward-noise standard deviations. Since all active benchmarks use diagonal $Q$, the proposal-whitening scale is computed elementwise as
\begin{align}
[\mathbf{s}_{\mathrm{white}}]_j = \max\!\left(c_{\mathrm{white}}[\boldsymbol{\sigma}_f]_j,\epsilon_{\mathrm{white}}\right), \qquad j=1,\ldots,d_x,
\end{align}
where $c_{\mathrm{white}}$ and $\epsilon_{\mathrm{white}}$ are fixed hyperparameters. In all experiments, $c_{\mathrm{white}}=1$ and $\epsilon_{\mathrm{white}}=10^{-3}$. Defining $D=\operatorname{diag}(\mathbf{s}_{\mathrm{white}})$, the whitened residual coordinate is
\begin{align}
\mathbf{z}_t = D^{-1}(\mathbf{x}_t-\hat{\mathbf{x}}_t).
\end{align}
This scaling normalizes proposal residuals to the scale of the forward uncertainty, thereby improving numerical conditioning during training.

The mean head predicts a residual mean in whitened coordinates. If $g^\mu_\theta(\mathbf{u}_t)$ denotes the raw mean output, the proposal mean in state coordinates is
\begin{align}
\boldsymbol{\mu}_\theta(\hat{\mathbf{x}}_t,\mathbf{y}_t,t) = \hat{\mathbf{x}}_t + D g^\mu_\theta(\mathbf{u}_t).
\label{eq:adx-residual-mean}
\end{align}
The covariance head predicts the $d_x(d_x+1)/2$ entries of a lower-triangular Cholesky factor $L_{\theta,z}$ in whitened coordinates. The corresponding Cholesky factor in state coordinates is
\begin{align}
L_{\theta,x} = D L_{\theta,z}.
\label{eq:adx-state-cholesky}
\end{align}
Positive definiteness is enforced through the Cholesky parameterization. The diagonal entries of $L_{\theta,z}$ are parameterized as
\begin{align}
[L_{\theta,z}]_{jj} = \operatorname{softplus}\!\left(\operatorname{clamp}(a_j,-5,20)\right),
\label{eq:adx-diagonal-first-three}
\end{align}
where $a_j$ is the corresponding raw diagonal output.

\paragraph{Residual network backbone.}
The default backbone in all experiments is a residual multilayer perceptron. The input layer maps $\mathbf{u}_t$ to a hidden representation $\mathbf{r}_0\in\mathbb{R}^H$. Each residual block computes
\begin{align}
\mathbf{u}_\ell &= \sigma(\operatorname{BN}_1(W_{1,\ell}\mathbf{r}_\ell+\mathbf{b}_{1,\ell})), \nonumber \\
\mathbf{v}_\ell &= \operatorname{BN}_2(W_{2,\ell}\mathbf{u}_\ell+\mathbf{b}_{2,\ell}), \\
\mathbf{r}_{\ell+1} &= \sigma(\mathbf{v}_\ell+\mathbf{r}_\ell), \nonumber
\end{align}
where $\sigma$ is ReLU and $\operatorname{BN}$ denotes batch normalization. Separate linear heads map the final hidden vector to the residual mean and Cholesky-vector outputs.

\begin{table}[H]
\centering
\caption{Neural proposal hyperparameters used in the experiments.}
\label{tab:adx-nn-hparams}
\begin{tabular}{lrrrrr}
\toprule
Benchmark & Hidden $H$ & Blocks &  Batch size & Training epochs \\
\midrule
Sensor bias system & 32 & 4 & 20 & 500 \\
Known-map localization & 64 & 4 &  50 & 500 \\
Range-only localization & 128 & 6 &  50 & 500 \\
3D unordered-range localization & 128 & 6 & 50 & 500 \\
\bottomrule
\end{tabular}
\end{table}

All proposal networks are trained using Adam with learning rate $10^{-3}$. The expectation with respect to $q_\theta$ in $\mathcal{L}_{\mathrm{ELBO}}(\theta;\mathbf{x}_{t-1},\mathbf{y}_t,t)$ is estimated using 100 Monte Carlo samples per training tuple $(\mathbf{x}_{t-1},\mathbf{y}_t,t)$ (see Section~\ref{adx:loss_function}).

\subsection{Loss Function}
\label{adx:loss_function}

For a fixed conditioning tuple $(\mathbf{x}_{t-1},\mathbf{y}_t,t)$, where dependence on $t$ is henceforth left implicit in the proposal parameterization, the optimal one-step proposal is $p(\mathbf{x}_t \mid \mathbf{x}_{t-1},\mathbf{y}_t)$. For any candidate proposal $q_\theta(\mathbf{x}_t \mid \mathbf{x}_{t-1},\mathbf{y}_t)$,
\begin{align}
D_{\mathrm{KL}}\!\left(q_\theta(\mathbf{x}_t \mid \mathbf{x}_{t-1},\mathbf{y}_t)\,\|\,p(\mathbf{x}_t \mid \mathbf{x}_{t-1},\mathbf{y}_t)\right)
=
\mathbb E_{q_\theta}\left[\log \frac{q_\theta(\mathbf{x}_t \mid \mathbf{x}_{t-1},\mathbf{y}_t)}{p(\mathbf{x}_t \mid \mathbf{x}_{t-1},\mathbf{y}_t)}\right].
\label{eq:adx-kl-start}
\end{align}

Using Bayes' rule,
\begin{align}
p(\mathbf{x}_t \mid \mathbf{x}_{t-1},\mathbf{y}_t)
=
\frac{p(\mathbf{y}_t \mid \mathbf{x}_t)p(\mathbf{x}_t \mid \mathbf{x}_{t-1})}{p(\mathbf{y}_t \mid \mathbf{x}_{t-1})},
\label{eq:adx-bayes-optimal-proposal}
\end{align}
which gives
\begin{align}
D_{\mathrm{KL}}(q_\theta\|p) =
&\mathbb E_{q_\theta}\left[\log q_\theta(\mathbf{x}_t \mid \mathbf{x}_{t-1},\mathbf{y}_t)-\log p(\mathbf{y}_t\mid \mathbf{x}_t)-\log p(\mathbf{x}_t\mid \mathbf{x}_{t-1})\right] \\ &+ \log p(\mathbf{y}_t\mid \mathbf{x}_{t-1}).
\label{eq:adx-kl-expanded}
\end{align}

Rearranging yields the ELBO identity
\begin{align}
\log p(\mathbf{y}_t\mid \mathbf{x}_{t-1})
=
&\mathbb E_{q_\theta}\left[\log p(\mathbf{y}_t\mid \mathbf{x}_t)+\log p(\mathbf{x}_t\mid \mathbf{x}_{t-1})-\log q_\theta(\mathbf{x}_t\mid \mathbf{x}_{t-1},\mathbf{y}_t)\right]\\ 
&+ D_{\mathrm{KL}}(q_\theta\|p).
\label{eq:adx-elbo-identity}
\end{align}

Since $\log p(\mathbf{y}_t\mid \mathbf{x}_{t-1})$ is constant with respect to $\theta$, minimizing the KL divergence is equivalent to maximizing the ELBO, or equivalently minimizing the negative ELBO loss used in Eq.~\ref{eqn:l_elbo}:
\begin{align}
\mathcal L_{\mathrm{ELBO}}(\theta;\mathbf{x}_{t-1},\mathbf{y}_t,t)
=
-\mathbb E_{q_\theta} \Big[\log p(\mathbf{y}_t\mid \mathbf{x}_t)+\log p(\mathbf{x}_t\mid \mathbf{x}_{t-1})-\log q_\theta(\mathbf{x}_t\mid \mathbf{x}_{t-1},\mathbf{y}_t) \Big].
\label{eq:adx-one-step-elbo}
\end{align}
The expectation with respect to $q_\theta$ is approximated by Monte Carlo using multiple samples drawn from $q_\theta$ for each conditioning tuple $(\mathbf{x}_{t-1},\mathbf{y}_t,t)$.

The full training objective is
\begin{align}
\mathcal L(\theta)
=
\mathbb E_{(\mathbf{x}_{t-1},\mathbf{y}_t,t)}
\left[
\mathcal L_{\mathrm{ELBO}}(\theta;\mathbf{x}_{t-1},\mathbf{y}_t,t)
\right],
\label{eq:adx-final-loss}
\end{align}
where the expectation over $(\mathbf{x}_{t-1},\mathbf{y}_t,t)$ is approximated by averaging $\mathcal L_{\mathrm{ELBO}}$ over minibatches of one-step transitions.

\subsection{Training and Inference Algorithms}
\label{adx:training_and_inference}

The training data is generated online during training; the data generation process is a simple rollout of the forward model

\begin{algorithm}[!htbp]
\caption{Training Data Generation via Simulated Trajectories}
\label{alg:data-generation}
\begin{algorithmic}[1]
\Require Forward model $\mathbf f$, Observation operator $\mathbf h$,
         initial distribution $p_0$, time grid $\{t_k\}_{k=0}^{T}$
\Require Number of trajectories $N_{\text{traj}} = 1000$

\For{$j = 1, \dots, N_{\text{traj}}$}
    \State $\mathbf{x}_0^{(j)} \sim p_0$ \Comment{Sample initial state from prior}
    \For{$k = 1, \dots, T$}
        \State $\mathbf{x}_k^{(j)} \leftarrow \mathbf{f}(\mathbf{x}_{k-1}^{(j)}, \mathbf{v}_k)$ \Comment{Stochastic Propagation}
        
        \State $\mathbf{y}_k^{(j)} \leftarrow \mathbf{h}(\mathbf{x}_k^{(j)}, \mathbf{w}_k)$ \Comment{Generate noisy observation}
    \EndFor

    \State \textbf{Return sliding-window pairs:}
    \State $\mathcal{D}^{(j)} \leftarrow \left\{ \left(\mathbf{x}_{k - 1}^{(j)},\; \mathbf{y}_{k}^{(j)},\; t_{k}\right) \right\}_{k=1}^{T}$
\EndFor
\State \Return $\mathcal{D} = \bigcup_{j=1}^{N_\text{traj}} \mathcal{D}^{(j)}$
\end{algorithmic}
\end{algorithm}

\begin{algorithm}[H]
\caption{Proposal Network Training for NOPFs}
\begin{algorithmic}[1]
\Require Neural proposal network $q_\theta(\mathbf{x}_t \mid \hat{\mathbf{x}}_t, \mathbf{y}_t, t)$ parameterized by $\theta$
\Require Forward model $\mathbf f$, Observation operator $\mathbf h$,
         initial distribution $p_0$, time grid $\{t_k\}_{k=0}^{T}$
\Require Number of MC samples $S$, learning rate $\eta$, batch size $B$

\State Initialize proposal network parameters $\theta$
\For{each epoch}
    \State Generate training dataset $\mathcal{D}$ via Algorithm~\ref{alg:data-generation}
    \For{each mini-batch $\{(\mathbf{x}_{t - 1}^{(b)},\; \mathbf{y}_{t}^{(b)},\; t^{(b)})\}_{b=1}^{B}$ from $\mathcal{D}$}
    \vspace{1mm}

    \For{each sample $b = 1, \dots, B$} 
        \State $\hat{\mathbf{x}}_{t}^{(b)} \leftarrow \mathbf{f}(\mathbf{x}_{t - 1}^{(b)})$ \Comment{Deterministic state propagation using noise-free model}
        \State $\boldsymbol{\mu}_\theta^{(b)},\; \text{vec}(\mathbf{L}_\theta^{(b)})
           \leftarrow \textsc{Net}_\theta\!\left(\hat{\mathbf{x}}^{(b)}_{t},\;
           \mathbf{y}_{t}^{(b)},\; t^{(b)}\right)$
        \State $\boldsymbol{\mu}_\theta^{(b)} \leftarrow \boldsymbol{\mu}_\theta^{(b)} + \hat{\mathbf{x}}^{(b)}$ \Comment{Residual connection}
        \State Construct $\mathbf{L}_\theta^{(b)}$ (lower-triangular) from $\text{vec}(\mathbf{L}_\theta^{(b)})$ \Comment{See Section~\ref{adx:proposal_parameterization}}
        \vspace{1mm}
        
        \For{$s = 1, \dots, S$} \Comment{Monte Carlo Estimate of $\mathcal{L}_\mathrm{ELBO}$}
            \State $\boldsymbol{\xi}^{(s)} \sim \mathcal{N}(\mathbf{0}, \mathbf{I}_{d_x})$
            \State $\mathbf{x}_{t}^{(b,s)} \leftarrow \boldsymbol{\mu}_\theta^{(b)}
                   + \mathbf{L}_\theta^{(b)}\,\boldsymbol{\xi}^{(s)}$
            \Comment{Reparameterization trick}
        \EndFor
        \vspace{1mm}
        \State Estimate $\mathcal L_{\mathrm{ELBO}}(\theta;\mathbf{x}_{t - 1}^{(b)},\; \mathbf{y}_{t}^{(b)},\; t^{(b)})$ using $S$ samples
    \EndFor
    \vspace{1mm}
    \State $\mathcal{L}(\theta) \leftarrow \frac{1}{B}
           \sum_{b=1}^{B} L_{\mathrm{ELBO}}(\theta;\mathbf{x}_{t - 1}^{(b)},\; \mathbf{y}_{t}^{(b)},\; t^{(b)})$ \Comment{Monte Carlo estimate over batch}
    \vspace{1mm}
    \State $\theta \leftarrow \theta - \eta \,\nabla_\theta \mathcal{L}(\theta)$
    
  \EndFor
\EndFor
\end{algorithmic}
\end{algorithm}

\begin{algorithm}[!htbp]
\caption{Inference with the Learned Particle Filter}
\begin{algorithmic}[1]
\Require Trained proposal $q_\theta$,Forward model $\mathbf f$, Observation operator $\mathbf h$, true observations $\{\mathbf{y}_k\}_{k=1}^{T}$, initial prior $p_0$, number of particles $N$
\Ensure Weighted particle approximation of the filtering distribution at each step

\State \textbf{Initialize:}\;
       $\{\mathbf{x}_0^{(i)}\}_{i=1}^{N} \sim p_0$,\quad
       $w_0^{(i)} \leftarrow 1/N$

\For{$k = 1, \dots, T$}
    \For{$i = 1, \dots, N$}
        \State $\hat{\mathbf{x}}_k^{(i)} \leftarrow f(\mathbf{x}_{k-1}^{(i)})$ \Comment{Deterministic state propagation using noise-free model}
        \State $\boldsymbol{\mu}^{(i)},\, \mathbf{L}^{(i)} \leftarrow q_\theta(\hat{\mathbf{x}}_k^{(i)},\, \mathbf{y}_k,\, t_k)$
        \State $\tilde{\mathbf{x}}_k^{(i)} \sim \mathcal{N}\!\left(\boldsymbol{\mu}^{(i)},\, \mathbf{L}^{(i)}{\mathbf{L}^{(i)}}^\top\right)$
        \State $\log \tilde{w}_k^{(i)} \leftarrow
           \log p(\mathbf{y}_k \mid \tilde{\mathbf{x}}_k^{(i)})
           + \log p(\tilde{\mathbf{x}}_k^{(i)} \mid \mathbf{x}_{k-1}^{(i)})
           - \log q_\theta(\tilde{\mathbf{x}}_k^{(i)} \mid \hat{\mathbf{x}}_k^{(i)},\, \mathbf{y}_k,\, t_k)$
           \vspace{2mm}
    \EndFor
    \vspace{2mm}
    \State $w_k^{(i)} \leftarrow
           \mathrm{softmax}\!\left(\{\log \tilde{w}_k^{(j)}\}_{j=1}^{N}\right)_i$ \Comment{Normalize Weights}

    \State $\{\mathbf{x}_k^{(i)},\, w_k^{(i)}\}
               \leftarrow \textsc{Resample}\!\left(
               \{\tilde{\mathbf{x}}_k^{(i)},\, w_k^{(i)}\}\right)$;\quad
               $w_k^{(i)} \leftarrow 1/N$ \Comment{Resample}
\EndFor

\State \textbf{Output:}\;
       $\Big\{ \left( \mathbf{x}_k^{(i)},\, w_k^{(i)} \right)_{i = 1}^N\Big\}_{k=1}^T$
\end{algorithmic}
\end{algorithm}

\section{Evaluation Metrics}
\label{adx:metrics_computation}

All particle metrics are reported as time averages over an evaluation index set $\mathcal{T} = \{1,\dots,T\}$, where $T$ is the trajectory length and
$\mathcal{T}$ excludes the initial transient (and, where applicable, the final
step at which a metric is undefined, for e.g., PNLL). The cardinality of this set is
$|\mathcal{T}|$. We denote the truth at time $t$ by $\mathbf{x}_t^\star$
(state) and $\mathbf{y}_t^\star$ (observation), the state dimension by $d_x$,
and the number of Monte Carlo replicates used by sampling-based estimators by
$M$.

\subsection{Normalized Mean Squared Error}
At every time step we form the weighted posterior mean
$\widehat{\boldsymbol{\mu}}_t$ using the linear average on non-angular
coordinates and the circular (atan2) average on angular coordinates.
The per-step mean squared error is the average squared component of the
wrapped residual between $\widehat{\boldsymbol{\mu}}_t$ and the truth,
\begin{align}
    \mathrm{MSE}_t = \tfrac{1}{d_x}\,\|\Delta_{\mathcal{A}}(\widehat{\boldsymbol{\mu}}_t,\mathbf{x}_t^\star)\|_2^2,
\end{align}
which is then time-averaged and normalized by the time-and-coordinate-averaged
squared truth on the same window:
\begin{equation}
\mathrm{NMSE}
=
\frac{
\dfrac{1}{|\mathcal{T}|}\sum_{t\in\mathcal{T}}\dfrac{1}{d_x}\,
\|\Delta_{\mathcal{A}}(\widehat{\boldsymbol{\mu}}_t,\mathbf{x}_t^\star)\|_2^2
}{
\dfrac{1}{|\mathcal{T}|\,d_x}\sum_{t\in\mathcal{T}}\|\mathbf{x}_t^\star\|_2^2
+\varepsilon
},
\label{eq:adx-nmse}
\end{equation}
where $\varepsilon$ is a small numerical floor (set to $10^{3}$ times machine
epsilon in the code) that prevents division by zero when the truth norm is
vanishingly small. NMSE is a point-estimate metric: it collapses the entire
particle cloud to its mean and is therefore not informative in multimodal
regimes where $\widehat{\boldsymbol{\mu}}_t$ falls in a region of low posterior
mass.

\subsection{Predictive Negative Log Likelihood}
The Predictive NLL (PNLL) scores how well the filter predicts the next
observation $\mathbf{y}_{t+1}^\star$ given everything observed up to time $t$.
Each particle is propagated one step through the stochastic dynamics (with randomly sample noise)
$\mathbf{x}_{t+1}^{(i)} = f(\mathbf{x}_t^{(i)},\mathbf{v}_t)$.
The one-step predictive density is approximated by the weighted particle
mixture
\begin{equation}
\widehat{p}(\mathbf{y}_{t+1}\mid\mathbf{y}_{1:t})
=
\sum_{i=1}^{N}
w_t^{(i)}\,
p\!\left(\mathbf{y}_{t+1}\,\big|\,\mathbf{x}_{t+1}^{(i)}\right),
\label{eq:adx-pnll-density}
\end{equation}
and the metric is the time-averaged negative log of this quantity evaluated at
the true next observation,
\begin{equation}
\mathrm{PNLL}
=
\frac{1}{|\mathcal{T}|}\sum_{t\in\mathcal{T}}
\Big[
-\log\widehat{p}(\mathbf{y}_{t+1}^\star\mid\mathbf{y}_{1:t}^\star)
\Big].
\label{eq:adx-pnll}
\end{equation}
The score is defined only for $t<T$, so the index $t=T$ is excluded from $\mathcal{T}$ for this metric. Unlike state-space scores, PNLL is well-defined even when the posterior is symmetric or label-ambiguous: any two states that produce the same observation distribution receive identical predictive scores.

\subsection{Energy Score}
The energy score, considering wrapping of the angular states, is defined as
\begin{align}
\mathrm{ES}_t = \mathbb{E}_{\mathbf{X}\sim\widehat{p}_t}[d_{\mathcal{A}}(\mathbf{X},\mathbf{x}_t^\star)] - \frac12 \mathbb{E}_{\mathbf{X},\mathbf{X}'\sim\widehat{p}_t}[d_{\mathcal{A}}(\mathbf{X},\mathbf{X}')].
\label{eqn:energy_score_adx}
\end{align}

The first expectation, $\mathbb{E}_{\mathbf{X}\sim\widehat{p}_t}[d_{\mathcal{A}}(\mathbf{X},\mathbf{x}_t^\star)]$, is computed under the weighted empirical distribution as $\sum_{i=1}^N w_t^{(i)}\,d_{\mathcal{A}}(\mathbf{x}_t^{(i)},\mathbf{x}_t^\star)$, since this requires only one wrapped-distance evaluation per particle.

The second expectation, $\mathbb{E}_{\mathbf{X},\mathbf{X}'\sim\widehat{p}_t}[d_{\mathcal{A}}(\mathbf{X},\mathbf{X}')]$, would otherwise be an $O(N^2)$ double sum and is replaced with an unbiased Monte Carlo estimator. We draw $2M$ particle indices $I_1,\dots,I_M,J_1,\dots,J_M$ i.i.d.\ from the categorical distribution $\mathrm{Cat}(w_t^{(1)},\dots,w_t^{(N)})$ (with replacement, using the same weights for both halves and resampled afresh at every $t$) and use
\begin{equation}
\widehat{\mathrm{ES}}_t = \sum_{i=1}^{N} w_t^{(i)}\,d_{\mathcal{A}}\!\left(\mathbf{x}_t^{(i)},\mathbf{x}_t^\star\right) - \frac{1}{2M}\sum_{m=1}^{M} d_{\mathcal{A}}\!\left(\mathbf{x}_t^{(I_m)},\mathbf{x}_t^{(J_m)}\right).
\label{eq:adx-es-empirical}
\end{equation}
The reported scalar is the time average $\mathrm{ES} = |\mathcal{T}|^{-1}\sum_{t\in\mathcal{T}} \widehat{\mathrm{ES}}_t$. The number of Monte Carlo pairs $M$ is a hyperparameter. We use $M=2048$, which makes the variance of the second term negligible relative to typical filter-to-filter differences. 


\subsection{Effective Sample Size}
The effective sample size (ESS) at time $t$ is reported in the standard
inverse-sum-of-squared-weights form,
\begin{equation}
\mathrm{ESS}_t = \left(\sum_{i=1}^{N}\bigl(w_t^{(i)}\bigr)^2\right)^{-1},
\label{eq:adx-ess}
\end{equation}
where $w_t^{(i)}$ are the normalized importance weights satisfying
$\sum_{i=1}^N w_t^{(i)} = 1$. By construction
$1 \le \mathrm{ESS}_t \le N$: the lower bound is reached when a single
particle carries unit weight (full degeneracy) and the upper bound when all
weights are uniform. The ESS is computed from the post-update weights
\emph{before} the systematic resampling step at each time index, so it
diagnoses weight degeneracy of the Sequential Importance Sampling step itself
rather than the residual Monte Carlo variance left after resampling. The
reported scalar is the time average
\begin{align}
   \mathrm{ESS} = \frac{1}{|\mathcal{T}|}\sum_{t\in\mathcal{T}} \mathrm{ESS}_t.
\end{align}
ESS does not by itself certify accuracy of the filter. It is purely a diagnostic of how informative the weighted ensemble is relative to the raw particle count, and is reported alongside accuracy metrics rather than in place of them.

\subsection{Wall Clock Time}
Wall clock time (WT) measures the end-to-end runtime, in seconds, of running one full filter on a single trajectory of length $T$ on CPU. Timing is recorded per seed and per filter and includes every operation invoked between the initialization of the particle ensemble at $t=0$ and the final state estimate at $t=T$. It includes proposal evaluation (including the forward pass through the learned proposal network for NOPF on CPU), state propagation via the dynamics, importance-weight updates, the systematic resampling step, and any per-step bookkeeping required by the specific filter variant.

It does \emph{not} include offline costs such as proposal training, data simulation, or post-hoc metric evaluation, which are amortized across runs and therefore not comparable on a per-trajectory basis. The reported scalar is the across-seed mean. To avoid hardware heterogeneity confounding the comparison, all filters in a given benchmark are timed on the same CPU node described in Appendix~\ref{adx:comp_arch}.

\subsection{Additional Metrics}
\label{adx:additional_metrics}

The additional metric reported includes Mean Absolute Deviation Error (MADE) and Joint Particle In Credible-region Probability (Joint PICP), which are presented for all benchmarks in Fig.~\ref{fig:additional_metrics} and are tabulated in Appendix~\ref{adx:tab_results_1}.

\begin{figure}[!htbp]
    \centering
    \includegraphics[width=\linewidth]{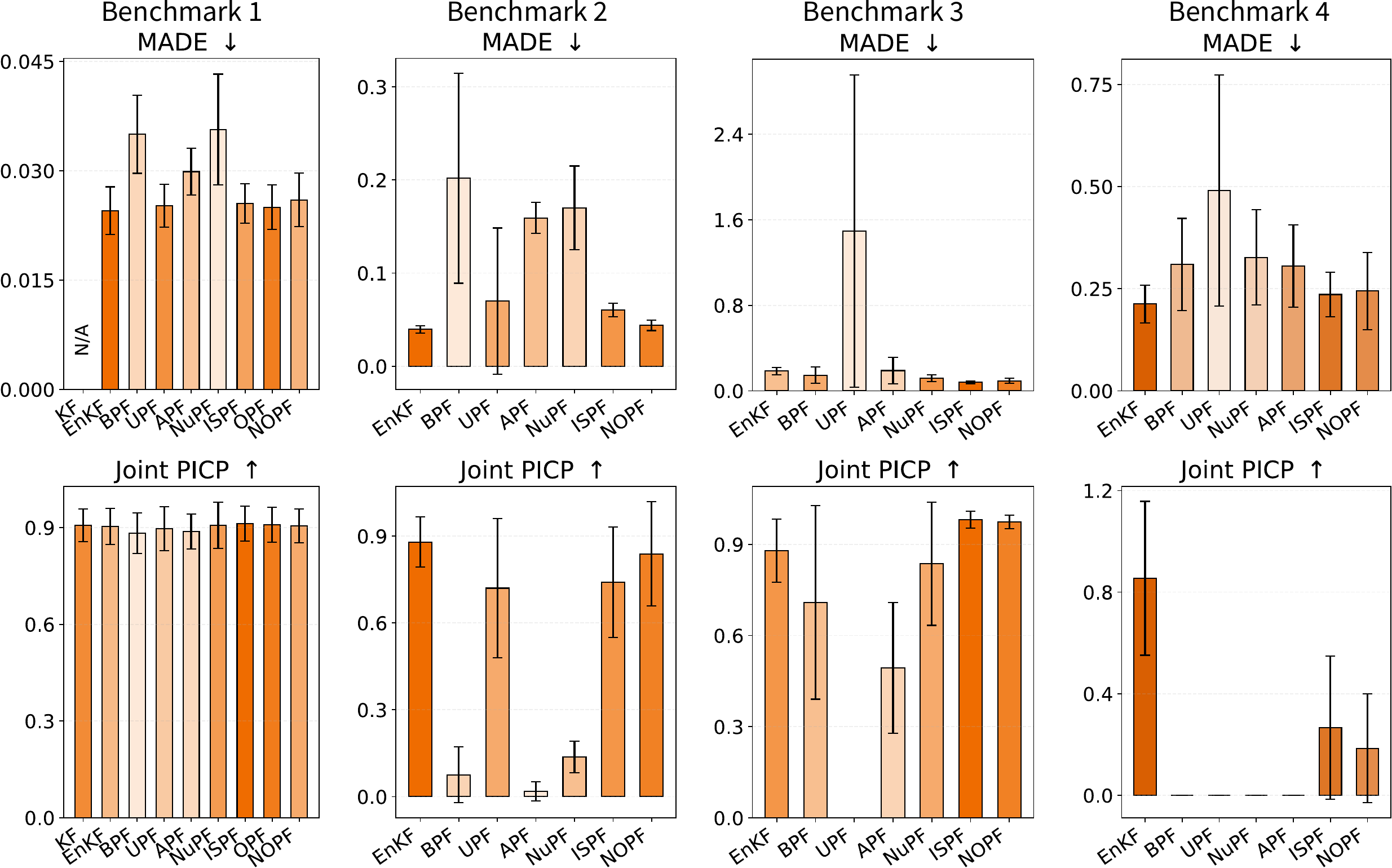}
    \caption{Additional metrics (MADE and Joint PICP) for benchmarks. Each column presents the metrics for a benchmark.}
    \label{fig:additional_metrics}
\end{figure}

\subsubsection{Mean Absolute Deviation Error}
MADE captures whether \emph{any} particle is close to the truth, regardless of its weight. At each time step we compute the wrapped product-metric distance between every particle and the true state and take the minimum, $\mathrm{MADE}_t = \min_{1\le i\le N}\,d_{\mathcal{A}}(\mathbf{x}_t^{(i)},\mathbf{x}_t^\star)$; the reported scalar is the time average
\begin{equation}
\mathrm{MADE} = \frac{1}{|\mathcal{T}|}\sum_{t\in\mathcal{T}} \min_{1\le i\le N}\, d_{\mathcal{A}}\!\left(\mathbf{x}_t^{(i)},\mathbf{x}_t^\star\right).
\label{eq:adx-made}
\end{equation}
Note that the importance weights $w_t^{(i)}$ do not enter Eq.~\eqref{eq:adx-made} and a single, possibly negligibly weighted, particle near the truth is sufficient to keep MADE small. MADE is therefore a support-coverage diagnostic complementary to NMSE: a filter whose ensemble collapses onto the wrong mode will have small MADE only if at least one surviving particle still tracks the truth.

\subsubsection{Joint Particle In Credible-region Probability}
The Joint PICP treats $\widehat{p}_t$ as a Gaussian surrogate $\mathcal{N}(\widehat{\boldsymbol{\mu}}_t, \widehat{\boldsymbol{\Sigma}}_t)$ on the full state and tests whether the truth lies inside the corresponding $(1-\alpha)$-coverage Mahalanobis ellipsoid. Concretely, we form weighted mean $\widehat{\boldsymbol{\mu}}_t$ and covariance $\widehat{\boldsymbol{\Sigma}}_t$ (a small isotropic ridge $10^{-8}I$ is added for numerical stability). We then evaluate the squared Mahalanobis distance of the truth from the weighted mean,
\begin{equation}
q_t = \Delta_{\mathcal{A}}(\mathbf{x}_t^\star,\widehat{\boldsymbol{\mu}}_t)^\top\, \widehat{\boldsymbol{\Sigma}}_t^{-1}\, \Delta_{\mathcal{A}}(\mathbf{x}_t^\star,\widehat{\boldsymbol{\mu}}_t),
\label{eq:adx-picp-q}
\end{equation}
and compare it to the $(1-\alpha)$ quantile of a chi-squared distribution with $d_x$ degrees of freedom, $\chi^2_{d_x}(1-\alpha)$. The metric is the empirical coverage rate
\begin{equation}
\mathrm{JointPICP}_\alpha = \frac{1}{|\mathcal{T}|}\sum_{t\in\mathcal{T}} \mathbf{1}\!\left[q_t \le \chi^2_{d_x}(1-\alpha)\right].
\label{eq:adx-joint-picp}
\end{equation}
We use $\alpha=0.1$, so a perfectly calibrated filter would yield $\mathrm{JointPICP}_{0.1}\approx 0.9$; values below indicate over-confident posteriors and values above indicate under-confident ones. The Gaussian surrogate is only an assumption for the calibration test, not for the underlying particle representation: $\widehat{p}_t$ itself remains a weighted empirical distribution, but the chi-squared threshold is what makes the per-step decision computable in $O(d_x^3)$ rather than via a kernel density tail integral. 

Because $q_t$ is compared against $\chi^2_{d_x}(1-\alpha)$, joint PICP is calibrated to the case in which the residual $\Delta_{\mathcal{A}}(\mathbf{x}_t^\star,\widehat{\boldsymbol{\mu}}_t)$ is Gaussian with covariance $\widehat{\boldsymbol{\Sigma}}_t$. When the underlying posterior departs from this assumption, the threshold is no longer the $(1-\alpha)$ quantile of $q_t$ and the metric becomes biased in either direction. Heavy-tailed, skewed, or weight-collapsed posteriors inflate $q_t$ above $\chi^2_{d_x}(1-\alpha)$ and cause systematic under-coverage, while strongly multimodal posteriors inflate $\widehat{\boldsymbol{\Sigma}}_t$ with the inter-mode spread and produce artificially high coverage even when the weighted mean lies in a low-density region between modes. Joint PICP should therefore be read as a calibration check on the Gaussian surrogate $\mathcal{N}(\widehat{\boldsymbol{\mu}}_t,\widehat{\boldsymbol{\Sigma}}_t)$ rather than on the true particle distribution $\widehat{p}_t$, and is best interpreted alongside shape-aware scores such as the energy score, which remain meaningful for non-Gaussian, multimodal, and heavy-tailed posteriors.

\section{Computing Environment}
\label{adx:comp_arch}

All proposal networks were trained on a Linux cluster using PyTorch on NVIDIA A100 80GB PCIe GPUs. Proposal training used a single GPU per benchmark together with CPU resources for data generation and batching. Propoasal were trained once and then the same learned proposal was used during inference across all seeds.

All inference filtering experiments and post-processing analyses were performed separately on a CPU node equipped with two Intel Xeon Silver 4216 processors (32 total cores; 16 cores per socket; 2.10 GHz base clock; two NUMA nodes), with one hardware thread per core. Independent random seeds were distributed across CPU workers for parallel experiment execution. All wall-clock runtimes (WT) reported in the main text, including forward passes through the learned proposal network, were measured on this CPU architecture to maintain consistent comparisons across particle filters.

Software was implemented in Python using PyTorch and NumPy under Linux. All neural proposal evaluations during filtering were executed on CPU for fair runtime comparison with competing particle filters. All forward-model and observation operator implementations are vectorized across particles.

\section{Particle Filter Implementations}
\label{adx:filter_details}

This section includes details of different filter implementations at time steps $t$ when the weighted ensemble $\{\mathbf{x}_{t-1}^{(i)}, \tilde{w}_{t-1}^{(i)}\}_{i=1}^N$ is provided as input.

\subsection{Bootstrap particle filter}

The bootstrap particle filter (BPF) samples
\begin{equation}
\mathbf{x}_t^{(i)} \sim p(\mathbf{x}_t \mid \mathbf{x}_{t-1}^{(i)})
\label{eq:adx-bpf-sample}
\end{equation}
and updates weights according to
\begin{equation}
\tilde{w}_t^{(i)}
=
w_{t-1}^{(i)} p(\mathbf{y}_t \mid \mathbf{x}_t^{(i)}),
\qquad
w_t^{(i)}
=
\frac{\tilde{w}_t^{(i)}}{\sum_{j=1}^N \tilde{w}_t^{(j)}}.
\label{eq:adx-bpf-weight_2}
\end{equation}

\subsection{Nudged particle filter}

NudgedPF first samples from the transition prior as
\begin{equation}
    \mathbf{x}_t^{(i)} \sim p(\mathbf{x}_t \mid \mathbf{x}_{t-1}^{(i)}),
\end{equation}
and then computes
\begin{equation}
    \mathbf{g}_t^{(i)} = \nabla_{\mathbf{x}_t} \log p(\mathbf{y}_t \mid \mathbf{x}_t) \big|_{\mathbf{x}_t = \mathbf{x}_t^{(i)}}.
\label{eq:adx-nudpf-grad}
\end{equation}
Computing the gradient of the log-likelihood is numerically more stable than evaluating the gradient of the likelihood, $\nabla_{\mathbf{x}_t} p(\mathbf{y}_t \mid \mathbf{x}_t)$, which can be very large or zero in flat regions.

Each particle is nudged independently with probability $N^{-1/2}$:
\begin{equation}
    \bar{\mathbf{x}}_t^{(i)} = \mathbf{x}_t^{(i)} + \eta I_t^{(i)} \mathbf{g}_t^{(i)}, \qquad I_t^{(i)} \sim \operatorname{Bernoulli}(N^{-1/2}).
\label{eq:adx-nudpf-nudge}
\end{equation}
The particles are then weighed using
\begin{equation}
\tilde{w}_t^{(i)} = w_{t-1}^{(i)} p(\mathbf{y}_t \mid \bar{\mathbf{x}}_t^{(i)}), \qquad w_t^{(i)} = \frac{\tilde{w}_t^{(i)}}{\sum_{j=1}^N \tilde{w}_t^{(j)}}.
\label{eq:adx-bpf-weight}
\end{equation} 
without adding a proposal correction term for the nudging transformation. The nudging strength parameter $\eta$ is selected as the highest $\eta$ for which the likelihood of most particles increases. The first benchmark uses $\eta = 10^{-2}$ while the rest of the benchmarks use $\eta = 2\times 10^{-4}$.

\subsection{Oracle Particle Filter}
\label{adx:oracle_pf}
The Oracle Particle Filter (OPF) is used only for the linear-Gaussian sensor-bias benchmark, where the locally optimal proposal is available in closed form. Consider the linear-Gaussian model
\begin{equation}
    \mathbf{x}_t = A\mathbf{x}_{t-1} + \boldsymbol{\epsilon}_t,
    \qquad
    \boldsymbol{\epsilon}_t \sim \mathcal{N}(0,Q),
\end{equation}
\begin{equation}
    \mathbf{y}_t = H\mathbf{x}_t + \boldsymbol{\eta}_t,
    \qquad
    \boldsymbol{\eta}_t \sim \mathcal{N}(0,R),
\end{equation}
where $A$ is the transition matrix, $H$ is the observation matrix, $Q$ is the process-noise covariance, and $R$ is the observation-noise covariance. For each ancestor particle $\mathbf{x}_{t-1}^{(i)}$, OraclePF samples from the locally optimal proposal
\begin{equation}
    q^\star(\mathbf{x}_t \mid \mathbf{x}_{t-1}^{(i)},\mathbf{y}_t)
    =
    p(\mathbf{x}_t \mid \mathbf{x}_{t-1}^{(i)},\mathbf{y}_t).
\end{equation}
This conditional distribution is Gaussian. Let
\begin{equation}
    S = HQH^\top + R,
    \qquad
    K = QH^\top S^{-1}.
\end{equation}
Then
\begin{equation}
    p(\mathbf{x}_t \mid \mathbf{x}_{t-1}^{(i)},\mathbf{y}_t)
    =
    \mathcal{N}
    \left(
    \mathbf{x}_t;
    \mathbf{m}_t^{(i)}, C
    \right),
\end{equation}
with
\begin{equation}
    \mathbf{m}_t^{(i)}
    =
    A\mathbf{x}_{t-1}^{(i)}
    +
    K
    \left(
    \mathbf{y}_t - HA\mathbf{x}_{t-1}^{(i)}
    \right),
    \qquad
    C
    =
    Q - KHQ.
\end{equation}
Equivalently, $C=(Q^{-1}+H^\top R^{-1}H)^{-1}$ and $\mathbf{m}_t^{(i)}=C(Q^{-1}A\mathbf{x}_{t-1}^{(i)}+H^\top R^{-1}\mathbf{y}_t)$. The Kalman-update form above is used in implementation for numerical stability.

After sampling $\mathbf{x}_t^{(i)} \sim \mathcal{N}(\mathbf{m}_t^{(i)},C)$, the outer weight is updated using the predictive likelihood
\begin{equation}
    p(\mathbf{y}_t \mid \mathbf{x}_{t-1}^{(i)})
    =
    \mathcal{N}
    \left(
    \mathbf{y}_t;
    HA\mathbf{x}_{t-1}^{(i)},
    S
    \right).
\end{equation}
Thus,
\begin{equation}
    \widetilde w_t^{(i)}
    =
    w_{t-1}^{(i)}
    p(\mathbf{y}_t \mid \mathbf{x}_{t-1}^{(i)}),
    \qquad
    w_t^{(i)}
    =
    \frac{\widetilde w_t^{(i)}}
    {\sum_{j=1}^N \widetilde w_t^{(j)}}.
\end{equation}

OPF provides a reference PF baseline for the linear-Gaussian benchmark, measuring how closely NPF approaches the closed-form locally optimal proposal in the setting where that proposal is available.

\subsection{Unscented particle filter}

UPF constructs a local Unscented Kalman Filter (UKF) proposal for each particle. The augmented state dimension for the UKF is
\begin{equation}
    d_{\mathrm{aug}} = 2d_x + d_y,
\label{eq:adx-upf-aug}
\end{equation}
corresponding to state, process noise, and observation noise. The sigma points constructed according to~\cite{wan2000unscented} with
\begin{equation}
    \lambda = \alpha^2(d_{\mathrm{aug}}+\kappa) - d_{\mathrm{aug}}.
\label{eq:adx-upf-lambda}
\end{equation}
induces a Gaussian proposal
$\mathcal{N}(\mathbf{m}_t^{(i)},P_t^{(i)})$ for each particle, and
\begin{equation}
\mathbf{x}_t^{(i)} \sim \mathcal{N}(\mathbf{m}_t^{(i)},P_t^{(i)}).
\label{eq:adx-upf-sample}
\end{equation}
The importance weight is evaluated as
\begin{equation}
    \tilde{w}_t^{(i)} = w_{t-1}^{(i)} \frac{ p(\mathbf{y}_t \mid \mathbf{x}_t^{(i)}) p(\mathbf{x}_t^{(i)} \mid \mathbf{x}_{t-1}^{(i)}) }{ q_{\mathrm{UKF}}(\mathbf{x}_t^{(i)} \mid \mathbf{x}_{t-1}^{(i)},\mathbf{y}_t)},
\label{eq:adx-upf-weight}
\end{equation}
followed by a normalization step.

The implementation adds diagonal jitter to the covariance matrix during sigma-point construction if the estimated covariance matrix is not positive definite. All UPFs use $\alpha = 10^{-3}$, $\beta = 2$, and $\kappa = 0$ for sigma point construction.

\subsection{Inner-Sampling Proposal Filter}
\label{adx:ispf}

The Inner-Sampling Proposal Filter (ISPF) is an expensive local proposal baseline that approximates the locally optimal proposal using inner Monte Carlo sampling. For each outer particle $\mathbf{x}_{t-1}^{(i)}$, we draw $M_{\mathrm{inner}}$ transition candidates
\begin{equation}
    \mathbf{x}_{t}^{(i,m)}
    \sim
    p(\mathbf{x}_t \mid \mathbf{x}_{t-1}^{(i)}),
    \qquad
    m = 1,\ldots,M_{\mathrm{inner}}.
\end{equation}
Each candidate is scored using the current likelihood,
\begin{equation}
    \ell_{t,m}^{(i)}
    =
    p(\mathbf{y}_t \mid \mathbf{x}_{t}^{(i,m)}),
\end{equation}
and normalized to obtain inner weights
\begin{equation}
    \alpha_{t,m}^{(i)}
    =
    \frac{\ell_{t,m}^{(i)}}
    {\sum_{r=1}^{M_{\mathrm{inner}}} \ell_{t,r}^{(i)}}.
\end{equation}
These weights define an empirical approximation to the local observation-informed proposal,
\begin{equation}
    \widehat q_t^{(i)}(d\mathbf{x})
    =
    \sum_{m=1}^{M_{\mathrm{inner}}}
    \alpha_{t,m}^{(i)}
    \delta_{\mathbf{x}_{t}^{(i,m)}}(d\mathbf{x}).
\end{equation}
We then draw an index
\begin{equation}
    K_i
    \sim
    \mathrm{Categorical}
    \left(
    \alpha_{t,1}^{(i)},\ldots,\alpha_{t,M_{\mathrm{inner}}}^{(i)}
    \right),
\end{equation}
and propagate the outer particle as
\begin{equation}
    \mathbf{x}_t^{(i)}
    =
    \mathbf{x}_{t}^{(i,K_i)}.
\end{equation}

The locally optimal proposal weight update requires the predictive likelihood
\begin{equation}
    p(\mathbf{y}_t \mid \mathbf{x}_{t-1}^{(i)})
    =
    \int
    p(\mathbf{y}_t \mid \mathbf{x}_t)
    p(\mathbf{x}_t \mid \mathbf{x}_{t-1}^{(i)})
    d\mathbf{x}_t.
\end{equation}
ISPF approximates this quantity using the same inner samples,
\begin{equation}
    \widehat Z_t^{(i)}
    =
    \frac{1}{M_{\mathrm{inner}}}
    \sum_{m=1}^{M_{\mathrm{inner}}}
    p(\mathbf{y}_t \mid \mathbf{x}_{t}^{(i,m)}).
\end{equation}
The outer weights are then updated as
\begin{equation}
    \widetilde w_t^{(i)}
    =
    w_{t-1}^{(i)}
    \widehat Z_t^{(i)},
    \qquad
    w_t^{(i)}
    =
    \frac{\widetilde w_t^{(i)}}
    {\sum_{j=1}^{N} \widetilde w_t^{(j)}}.
\end{equation}
The selected candidate likelihood is not applied again in the outer weight update, since the inner categorical draw already samples candidates with probability proportional to their likelihood. Thus, ISPF is a stochastic inner-sampling approximation to the locally optimal proposal. In all experiments, we use $M_{\mathrm{inner}}=200$ inner samples per outer particle.

\subsection{Auxiliary particle filter}

The auxiliary particle filter (APF) introduces an ancestor-selection stage that uses the current observation $\mathbf{y}_t$ to focus subsequent propagation on particles likely to explain it. Each particle is first advanced to its deterministic predictive mean
\begin{equation}
\hat{\mathbf{x}}_t^{(i)} = \bar{\mathbf{f}}(\mathbf{x}_{t-1}^{(i)}),
\label{eq:adx-apf-predictive-mean}
\end{equation}
and assigned a first-stage (auxiliary) weight that combines the prior normalized weight with the observation likelihood evaluated at this mean,
\begin{equation}
\lambda_t^{(i)} \propto w_{t-1}^{(i)}\,p(\mathbf{y}_t \mid \hat{\mathbf{x}}_t^{(i)}),
\qquad
\sum_{i=1}^N \lambda_t^{(i)} = 1.
\label{eq:adx-apf-first-stage}
\end{equation}
Ancestor indices $a_i \in \{1,\dots,N\}$ are drawn with replacement from $\operatorname{Categorical}(\lambda_t^{(1)},\dots,\lambda_t^{(N)})$, and the selected ancestors are propagated through the full stochastic forward map,
\begin{equation}
\mathbf{x}_t^{(i)} = \mathbf{f}(\mathbf{x}_{t-1}^{a_i},\mathbf{v}_t)
\label{eq:adx-apf-prop}
\end{equation}
The second-stage weight removes the bias introduced by the auxiliary draw using the ratio of the true likelihood at the propagated particle to the predictive likelihood used to select its ancestor
\begin{equation}
\tilde{w}_t^{(i)} \propto \frac{p(\mathbf{y}_t \mid \mathbf{x}_t^{(i)})}{p(\mathbf{y}_t \mid \hat{\mathbf{x}}_t^{a_i})},
\qquad
w_t^{(i)} = \frac{\tilde{w}_t^{(i)}}{\sum_{j=1}^N \tilde{w}_t^{(j)}}.
\label{eq:adx-apf-second-stage}
\end{equation}

\subsection{Ensemble Kalman filter}

The Ensemble Kalman Filter (EnKF) replaces the analytic covariance in the Kalman update with empirical estimates obtained from a Monte Carlo ensemble. At time $t$, each ensemble member is propagated through the stochastic forward model and pushed through the observation operator with an independent observation-noise sample to produce a paired state-observation ensemble,
\begin{equation}
\mathbf{x}_t^{(i)} = \mathbf{f}(\mathbf{x}_{t-1}^{(i)},\mathbf{v}_t),
\qquad
\mathbf{y}_t^{(i)} = \mathbf{h}(\mathbf{x}_t^{(i)}, \mathbf{e}_t).
\label{eq:adx-enkf-prop}
\end{equation}
The empirical means $\bar{\mathbf{x}}_t = \tfrac{1}{N}\sum_i \mathbf{x}_t^{(i)}$ and $\bar{\mathbf{y}}_t = \tfrac{1}{N}\sum_i \mathbf{y}_t^{(i)}$ define the centered deviations $\delta\mathbf{x}_t^{(i)} = \mathbf{x}_t^{(i)} - \bar{\mathbf{x}}_t$ and $\delta\mathbf{y}_t^{(i)} = \mathbf{y}_t^{(i)} - \bar{\mathbf{y}}_t$, from which the cross-covariance and observation-covariance are estimated as
\begin{equation}
P_{xy} = \frac{1}{N}\sum_{i=1}^N \delta\mathbf{x}_t^{(i)} (\delta\mathbf{y}_t^{(i)})^\top,
\qquad
P_{yy} = \frac{1}{N}\sum_{i=1}^N \delta\mathbf{y}_t^{(i)} (\delta\mathbf{y}_t^{(i)})^\top.
\label{eq:adx-enkf-cov}
\end{equation}
The ensemble Kalman gain is $K_t = P_{xy} P_{yy}^{-1}$, and each member is corrected toward the true observation $\mathbf{y}_t$ using its own synthetic observation $\mathbf{y}_t^{(i)}$ as a stochastic innovation,
\begin{equation}
\mathbf{x}_t^{(i)} = \hat{\mathbf{x}}_t^{(i)} + K_t(\mathbf{y}_t - \hat{\mathbf{y}}_t^{(i)}),
\label{eq:adx-enkf-update}
\end{equation}
where $\hat{\mathbf{x}}_t^{(i)} = \bar{\mathbf{f}}(\mathbf{x}_{t-1}^{(i)})$ and $\hat{\mathbf{y}}_t^{(i)} = \bar{\mathbf{h}}(\hat{\mathbf{y}}_t^{(i)})$. The updated ensemble $\{\mathbf{x}_t^{(i)}\}_{i=1}^N$ carries no importance weights, and for downstream metric computation, each member is assigned a uniform weight $w_t^{(i)} = 1/N$. EnKF recovers the Kalman update for the filtering distribution $p(\mathbf{x}_t \mid \mathbf{y}_{1:t})$ when $(\mathbf{x}_t,\mathbf{y}_t)$ are jointly Gaussian in the large-ensemble limit, and remains robust and widely used in practice.

\subsection{Neural Optimal Particle filter}

The Neural Optimal Particle Filter (NOPF) replaces the importance density with a learned conditional distribution $q_\theta(\mathbf{x}_t \mid \mathbf{x}_{t-1},\mathbf{y}_t,t)$ trained as described in Appendix~\ref{adx:loss_function}. To keep the network from having to relearn the dynamics, each particle is first advanced to its deterministic predictive mean
\begin{equation}
\hat{\mathbf{x}}_t^{(i)} = \Bar{\mathbf{f}}(\mathbf{x}_{t-1}^{(i)}),
\label{eq:adx-NOPF-predict}
\end{equation}
and $\hat{\mathbf{x}}_t^{(i)}$ is supplied to the proposal network as an auxiliary conditioning input so the network only needs to model the observation-informed correction to the deterministic skeleton. A new particle is then drawn from the learned proposal,
\begin{equation}
\mathbf{x}_t^{(i)} \sim q_\theta(\mathbf{x}_t \mid \mathbf{x}_{t-1}^{(i)},\mathbf{y}_t,t),
\label{eq:adx-NOPF-sample}
\end{equation}
and the standard SIS weight is evaluated in log-space using the observation likelihood, the transition density, and the proposal density,
\begin{equation}
    \tilde{w}_t^{(i)} = w_{t-1}^{(i)} \frac{ p(\mathbf{y}_t \mid \mathbf{x}_t^{(i)}) p(\mathbf{x}_t^{(i)} \mid \mathbf{x}_{t-1}^{(i)}) }{ q_{\theta}(\mathbf{x}_t^{(i)} \mid \mathbf{x}_{t-1}^{(i)},\mathbf{y}_t, t)},
\label{eq:adx-NOPF-weight}
\end{equation}
The forward pass through $q_\theta$ adds one network evaluation per particle on top of the cost of a bootstrap step, so the per-step cost is $\mathcal{O}(N\,c_\theta)$ where $c_\theta$ is the network cost of propagating a single particle.

\subsection{Systematic resampling}

Particle filters use systematic resampling. Given normalized weights
$w_t^{(i)}$, sample $u_0 \sim \operatorname{Uniform}(0,1/N)$ and define
\begin{equation}
u_i = u_0 + \frac{i-1}{N},
\qquad i=1,\ldots,N.
\label{eq:adx-systematic-grid}
\end{equation}
The $i$th resampled particle is the first one whose cumulative weight exceeds
$u_i$. After resampling, all weights are reset to $1/N$.

\subsection{Computational Complexity of PFs}
\label{adx:comp_complexity}

Let $d_x$ and $d_y$ denote the dimensions of the state and observation spaces, respectively, and let $n_p$ denote the number of outer particles in a particle filter. For ISPF, $n_q$ denotes the number of inner samples per outer particle. The cost of one evaluation of the forward model $\mathbf{f}$ and the observation operator $\mathbf{h}$ (Eq.~\ref{eqn:problem_setup}) is denoted by $\mathcal{O}(\mathbf{f})$ and $\mathcal{O}(\mathbf{h})$, and the cost of a single forward pass through the proposal network is $\mathcal{O}(nn)$. Per-step asymptotic costs are summarized in Table~\ref{tab:CompComplexity}; systematic resampling, Gaussian density evaluations, and other operations linear in $n_p$ are absorbed into the listed terms.

\begin{table}[!htbp]
\centering
\setlength{\tabcolsep}{4pt}
\renewcommand{\arraystretch}{1.4}
\resizebox{\textwidth}{!}{%
\begin{tabular}{|l|c|c|c|c|}
\hline
\textbf{Filter} & \textbf{Propagation} & \textbf{Importance Sampling} & \textbf{Weight Update} & \textbf{Total} \\
\hline
BPF
& $n_p\mathcal{O}(\mathbf{f})$
& --
& $n_p\mathcal{O}(\mathbf{h})$
& $n_p(\mathcal{O}(\mathbf{f}) + \mathcal{O}(\mathbf{h}))$ \\
\hline
NuPF
& $n_p\mathcal{O}(\mathbf{f})$
& $n_p\mathcal{O}(\mathbf{h})$
& $n_p\mathcal{O}(\mathbf{h})$
& $n_p\mathcal{O}(\mathbf{f}) + 2n_p\mathcal{O}(\mathbf{h})$ \\
\hline
UPF
& \multicolumn{2}{c|}{$n_p(4d_x+2d_y+1)(\mathcal{O}(\mathbf{f}) + \mathcal{O}(\mathbf{h})) + n_p\mathcal{O}((2d_x+d_y)^3)$}
& $n_p\mathcal{O}(\mathbf{h})$
& $n_p(4d_x+2d_y+1)\mathcal{O}(\mathbf{f}) + n_p(4d_x+2d_y+2)\mathcal{O}(\mathbf{h}) + n_p\mathcal{O}((2d_x+d_y)^3)$ \\
\hline
ISPF
& \multicolumn{2}{c|}{$n_pn_q(\mathcal{O}(\mathbf{f}) + \mathcal{O}(\mathbf{h}))$}
& 
& $n_pn_q(\mathcal{O}(\mathbf{f}) + \mathcal{O}(\mathbf{h})) $ \\
\hline
APF
& $2n_p\mathcal{O}(\mathbf{f})$
& $n_p\mathcal{O}(\mathbf{h})$
& $n_p\mathcal{O}(\mathbf{h})$
& $2n_p\mathcal{O}(\mathbf{f}) + 2n_p\mathcal{O}(\mathbf{h})$ \\
\hline
EnKF
& \multicolumn{3}{c|}{$n_p(\mathcal{O}(\mathbf{f}) + \mathcal{O}(\mathbf{h})) + \mathcal{O}(n_pd_xd_y + n_pd_y^2 + d_y^3)$}
& $n_p(\mathcal{O}(\mathbf{f}) + \mathcal{O}(\mathbf{h})) + \mathcal{O}(n_pd_xd_y + n_pd_y^2 + d_y^3)$ \\
\hline
NOPF
& $n_p\mathcal{O}(\mathbf{f})$
& $n_p\mathcal{O}(nn)$
& $n_p\mathcal{O}(\mathbf{h})$
& $n_p(\mathcal{O}(\mathbf{f}) + \mathcal{O}(\mathbf{h})) + n_p\mathcal{O}(nn)$ \\
\hline
\end{tabular}%
}
\caption{Per-step computational complexity of the particle filters considered in this work.}
\label{tab:CompComplexity}
\end{table}

We assume additive process noise, so the transition density $p(\mathbf{x}_t \mid \mathbf{x}_{t-1})$ can be evaluated from the deterministic forward map cached during the propagation step without an additional $\mathbf{f}$ call.

\section{Benchmark Specifications}
\label{adx:experiment_details}

Each benchmark specializes the state-space model in Eq.~\eqref{eqn:problem_setup} with additive process noise $\mathbf v_t\sim\mathcal N(\mathbf 0,\mathbf Q)$, where $\mathbf Q$ is diagonal. For the first three benchmarks, the observation noise is additive with diagonal covariance $\mathbf R$, so $\mathbf y_t=\bar{\mathbf h}(\mathbf x_t)+\mathbf w_t$. For the current positive-prior 3D unordered-range benchmark, the proprioceptive channels are additive, but independent range noise is applied to the four labeled raw ranges before sorting; the likelihood is therefore written as a permutation-marginalized unordered-range likelihood in Appendix~\ref{adx:benchmark_4}. We write $\bar{\mathbf f}$ for the deterministic skeleton of $\mathbf f$ (i.e.\ $\bar{\mathbf f}(\mathbf x_{t-1})=\mathbf f(\mathbf x_{t-1},\mathbf 0)$). All angular dimensions are wrapped to $(-\pi,\pi]$. Throughout, $\boldsymbol\sigma_f$ and $\boldsymbol\sigma_h$ denote per-coordinate process and observation standard deviations when those diagonal covariances are defined. Table~\ref{tab:adx-benchmark-overview} summarizes the key parameters for the four benchmarks.

\begin{table}[H]
\centering
\caption{Benchmark details.}
\label{tab:adx-benchmark-overview}
\begin{tabular}{lrrrr}
\toprule
Benchmark & $d_x$ & $d_y$ & Time points & Particles \\
\midrule
Sensor bias system & 3 & 2 & 60 & 2000 \\
Known-map localization & 7 & 10 & 51 & 10000 \\
Range-only localization & 8 & 3 & 60 & 10000 \\
3D unordered-range localization & 20 & 4 & 35 & 50000 \\
\bottomrule
\end{tabular}
\end{table}

\subsection{Benchmark 1: Linear System with Biased Sensors}
\label{adx:benchmark_1}
The sensor-bias benchmark has state
\begin{equation}
\mathbf x_t=[p_t,\,v_t,\,b_t]^\top\in\mathbb R^3,
\label{eq:adx-sbs-state}
\end{equation}
where $p_t$ is position, $v_t$ velocity, and $b_t$ an additive sensor bias. The time grid is $t_k=0.1(k+1)$ for $k=0,\ldots,59$. Specializing Eq.~\eqref{eqn:problem_setup} to a linear-Gaussian state-space model gives
\begin{equation}
\mathbf x_t=\mathbf A\mathbf x_{t-1}+\mathbf v_t,
\qquad
\mathbf y_t=\mathbf H\mathbf x_t+\mathbf w_t,
\label{eq:adx-sbs-model}
\end{equation}
with
\begin{equation}
\mathbf A=\begin{bmatrix}1&\Delta t&0\\0&1&0\\0&0&1\end{bmatrix},\qquad
\mathbf H=\begin{bmatrix}1&0&0\\0&1&1\end{bmatrix},\qquad
\Delta t=0.1,
\label{eq:adx-sbs-AH}
\end{equation}
and diagonal noise covariances
\begin{equation}
\mathbf Q=\diag(0.1^2,\,0.1^2,\,0.1^2),
\quad
\mathbf R=\diag(0.1^2,\,0.1^2),
\quad
\mathbf v_t\sim\mathcal N(\mathbf 0,\mathbf Q),\ \mathbf w_t\sim\mathcal N(\mathbf 0,\mathbf R).
\label{eq:adx-sbs-noise}
\end{equation}
The prior is
\begin{equation}
p_0(\mathbf x_0)=\mathcal N(\boldsymbol\mu_0,\boldsymbol\Sigma_0),
\qquad
\boldsymbol\mu_0=[0.8,\,0.2,\,-0.1]^\top,
\qquad
\boldsymbol\Sigma_0=\diag(0.5,\,0.5,\,0.5).
\label{eq:adx-sbs-prior}
\end{equation}

\subsection{Benchmark 2: 2D Robot Localization with Range and Bearing Observations}
\label{adx:benchmark_2}

The known-map localization benchmark estimates robot pose, motion states, and odometry biases with a fixed landmark map. The state is
\begin{equation}
\mathbf x_t=[p_t^x,\,p_t^y,\,\theta_t,\,v_t,\,\omega_t,\,b_t^v,\,b_t^\omega]^\top\in\mathbb R^7,
\label{eq:adx-loc-state}
\end{equation}
with $(p_t^x,p_t^y)$ position, $\theta_t$ heading, $v_t$ forward speed, $\omega_t$ yaw rate, and $(b_t^v,b_t^\omega)$ slowly varying odometry biases. The four landmarks are
\begin{equation}
\boldsymbol\ell_1=(-4.5,1.8),\quad
\boldsymbol\ell_2=(3.8,3.1),\quad
\boldsymbol\ell_3=(-2.2,-3.8),\quad
\boldsymbol\ell_4=(4.4,-2.6),
\label{eq:adx-loc-landmarks}
\end{equation}
and the time grid is $t_k=0.1k$ for $k=0,\ldots,50$ with $\Delta t=0.1$, $\rho_v=2.0$, $\rho_\omega=2.7$. Specializing Eq.~\eqref{eqn:problem_setup} with additive process noise gives $\mathbf x_t=\bar{\mathbf f}(\mathbf x_{t-1},t-1)+\mathbf v_t$, where $\bar{\mathbf f}$ is defined component-wise by
\begin{align}
p_t^x&=p_{t-1}^x+\Delta t\,v_{t-1}\cos\theta_{t-1},
&
p_t^y&=p_{t-1}^y+\Delta t\,v_{t-1}\sin\theta_{t-1}, \nonumber\\
\theta_t&=\wrap(\theta_{t-1}+\Delta t\,\omega_{t-1}),
&
v_t&=v_{t-1}+\rho_v\bigl(v_{\mathrm{cmd}}(t-1)-v_{t-1}\bigr)\Delta t, \nonumber\\
\omega_t&=\omega_{t-1}+\rho_\omega\bigl(\omega_{\mathrm{cmd}}(t-1)-\omega_{t-1}\bigr)\Delta t,
&
b_t^v&=b_{t-1}^v,\qquad b_t^\omega=b_{t-1}^\omega,
\label{eq:adx-loc-dynamics}
\end{align}
so the bias states evolve as a Gaussian random walk through the additive process noise term $\mathbf v_t$. The command schedule is
\begin{align}
v_{\mathrm{cmd}}(t)&=0.85+0.20\sin(0.33t)+0.08\cos(0.07t),\\
\omega_{\mathrm{cmd}}(t)&=0.02+0.42\sin(0.21t)+0.12\cos(0.09t).
\end{align}
Process noise is $\mathbf v_t\sim\mathcal N(\mathbf 0,\mathbf Q)$ with $\mathbf Q=\diag(\boldsymbol\sigma_f^2)$ and
\begin{equation}
\boldsymbol\sigma_f=[0.20,\,0.20,\,1^\circ,\,0.40,\,4^\circ,\,0.30,\,0.20^\circ]^\top.
\label{eq:adx-loc-process-noise}
\end{equation}
For each landmark $\boldsymbol\ell_j=(\ell_j^x,\ell_j^y)$, define the relative offsets, range, and bearing
\begin{align}
&\delta_{j,t}^x=\ell_j^x-p_t^x,\quad
\delta_{j,t}^y=\ell_j^y-p_t^y, \\ \nonumber
&r_{j,t}=\sqrt{(\delta_{j,t}^x)^2+(\delta_{j,t}^y)^2},\quad
\beta_{j,t}=\wrap\!\bigl(\operatorname{atan2}(\delta_{j,t}^y,\delta_{j,t}^x)-\theta_t\bigr).
\label{eq:adx-loc-relative}
\end{align}
The observation $\mathbf y_t=\bar{\mathbf h}(\mathbf x_t)+\mathbf w_t$ stacks the four range-bearing pairs and the two biased odometry channels,
\begin{equation}
\bar{\mathbf h}(\mathbf x_t)=
[r_{1,t},\,\beta_{1,t},\,r_{2,t},\,\beta_{2,t},\,r_{3,t},\,\beta_{3,t},\,r_{4,t},\,\beta_{4,t},\,v_t+b_t^v,\,\omega_t+b_t^\omega]^\top\in\mathbb R^{10},
\label{eq:adx-loc-observation}
\end{equation}
and $\mathbf w_t\sim\mathcal N(\mathbf 0,\mathbf R)$ with $\mathbf R=\diag(\boldsymbol\sigma_h^2)$ and
\begin{equation}
\boldsymbol\sigma_h=[\,\underbrace{0.08,\,1.5^\circ,\,0.08,\,1.5^\circ,\,0.08,\,1.5^\circ,\,0.08,\,1.5^\circ}_{\text{range, bearing per landmark}},\;0.05,\,1.2^\circ\,]^\top.
\label{eq:adx-loc-obs-noise}
\end{equation}
The initial distribution is Gaussian, $p_0(\mathbf x_0)=\mathcal N(\boldsymbol\mu_0,\boldsymbol\Sigma_0)$, with
\begin{align}
\boldsymbol\mu_0=[-3.0,\,-2.0,\,0.3,\,0.7,\,0.0,\,0.08,\,-0.05]^\top,
\\ \nonumber
\boldsymbol\Sigma_0=\diag\!\bigl(0.40^2,\,0.40^2,\,(10^\circ)^2,\,0.12^2,\,(6^\circ)^2,\,0.05^2,\,(2^\circ)^2\bigr).
\label{eq:adx-loc-prior}
\end{align}

\subsection{Benchmark 3: 2D Robot Localization with Biased Range-Only Observations}
\label{adx:benchmark_3}
The range-only benchmark has state
\begin{equation}
\mathbf x_t=[p_t^x,\,p_t^y,\,\theta_t,\,v_t,\,\omega_t,\,b_t^r,\,b_t^v,\,b_t^\omega]^\top\in\mathbb R^8,
\label{eq:adx-rol-state}
\end{equation}
with anchor $\mathbf a=(0,0)$, $\Delta t=0.1$, and $t_k=0.1k$ for $k=0,\ldots,59$. Specializing Eq.~\eqref{eqn:problem_setup} with additive process noise gives $\mathbf x_t=\bar{\mathbf f}(\mathbf x_{t-1},t-1)+\mathbf v_t$, with
\begin{align}
p_t^x&=p_{t-1}^x+\Delta t\,v_{t-1}\cos\theta_{t-1},
&
p_t^y&=p_{t-1}^y+\Delta t\,v_{t-1}\sin\theta_{t-1}, \nonumber\\
\theta_t&=\wrap(\theta_{t-1}+\Delta t\,\omega_{t-1}),
&
v_t&=v_{t-1},\qquad \omega_t=\omega_{t-1}, \nonumber\\
b_t^r&=b_{t-1}^r,
&
b_t^v&=b_{t-1}^v,\qquad b_t^\omega=b_{t-1}^\omega.
\label{eq:adx-rol-dynamics}
\end{align}
All bias and motion states therefore evolve as Gaussian random walks once the process noise $\mathbf v_t$ is added, with $\mathbf v_t\sim\mathcal N(\mathbf 0,\mathbf Q)$, $\mathbf Q=\diag(\boldsymbol\sigma_f^2)$, and
\begin{equation}
\boldsymbol\sigma_f=[0.035,\,0.035,\,0.8^\circ,\,0.035,\,1.2^\circ,\,0.012,\,0.012,\,0.2^\circ]^\top.
\label{eq:adx-rol-process-noise}
\end{equation}
The observation is $\mathbf y_t=\bar{\mathbf h}(\mathbf x_t)+\mathbf w_t$ with
\begin{equation}
\bar{\mathbf h}(\mathbf x_t)=
\bigl[\,\|\mathbf p_t-\mathbf a\|_2+b_t^r,\;v_t+b_t^v,\;\omega_t+b_t^\omega\,\bigr]^\top,
\qquad
\mathbf p_t=[p_t^x,p_t^y]^\top,
\label{eq:adx-rol-observation}
\end{equation}
and $\mathbf w_t\sim\mathcal N(\mathbf 0,\mathbf R)$, $\mathbf R=\diag(\boldsymbol\sigma_h^2)$,
\begin{equation}
\boldsymbol\sigma_h=[0.025,\,0.04,\,1^\circ]^\top.
\label{eq:adx-rol-obs-noise}
\end{equation}
The filtering prior is a mirrored two-component Gaussian mixture,
\begin{equation}
k\sim\Cat(0.5,0.5),\qquad
\mathbf x_0\,|\,k\sim\mathcal N\!\bigl(\boldsymbol\mu_0^{(k)},\boldsymbol\Sigma_0^{\mathrm{mode}}\bigr),
\label{eq:adx-rol-filtering-prior-cat}
\end{equation}
with
\begin{align}
\boldsymbol\mu_0^{(1)}&=[3.0,\,0.6,\,\pi,\,0.42,\,0.02,\,0.0,\,0.0,\,0.0]^\top,\\
\boldsymbol\mu_0^{(2)}&=[-3.0,\,0.6,\,\pi,\,0.42,\,0.02,\,0.0,\,0.0,\,0.0]^\top,\\
\boldsymbol\Sigma_0^{\mathrm{mode}}&=\diag\!\bigl(0.70^2,\,0.25^2,\,(8^\circ)^2,\,0.16^2,\,(4^\circ)^2,\,0.40^2,\,0.12^2,\,(2^\circ)^2\bigr).
\label{eq:adx-rol-filtering-prior}
\end{align}

\subsection{Benchmark 4: 3D Unordered-Range Localization}
\label{adx:benchmark_4}
The 3D unordered-range benchmark has state
\begin{equation}
\begin{aligned}
\mathbf x_t=\big[\;
&p_t^x,\,p_t^y,\,p_t^z,\;\psi_t,\,\theta_t,\;v_t,\,\omega_t^\psi,\,\omega_t^\theta,\;
b_t^v,\,b_t^{\omega^\psi},\,b_t^{\omega^\theta},\\
&b_t^{r_1},\,b_t^{r_2},\,b_t^{r_3},\,b_t^{r_4},\;
d_t^x,\,d_t^y,\,d_t^z,\;s_t^v,\,s_t^r\;\big]^\top\in\mathbb R^{20},
\end{aligned}
\label{eq:adx-uro20d-state}
\end{equation}
where $\mathbf p_t=[p_t^x,p_t^y,p_t^z]^\top$ is position, $(\psi_t,\theta_t)$ are yaw and pitch, $v_t$ is forward speed, $(\omega_t^\psi,\omega_t^\theta)$ are angular rates, $(b_t^v,b_t^{\omega^\psi},b_t^{\omega^\theta})$ are proprioceptive biases, $(b_t^{r_1},\ldots,b_t^{r_4})$ are per-anchor range biases, $(d_t^x,d_t^y,d_t^z)$ are slowly varying drift disturbances on position, and $(s_t^v,s_t^r)$ are multiplicative scale-calibration factors on speed and range. The four range anchors are
\begin{equation}
\mathbf a_1=(-4.4,-2.1,0),\quad
\mathbf a_2=(4.2,-1.3,0),\quad
\mathbf a_3=(-0.8,3.8,0),\quad
\mathbf a_4=(4.7,3.4,0).
\label{eq:adx-uro20d-anchors}
\end{equation}
The time grid is $t_k=0.1k$ for $k=0,\ldots,34$, with $\Delta t=0.1$, $\rho_v=1.7$, $\rho_\psi=2.1$, and $\rho_\theta=2.3$.. Specializing Eq.~\eqref{eqn:problem_setup} with additive process noise gives $\mathbf x_t=\bar{\mathbf f}(\mathbf x_{t-1})+\mathbf v_t$. The pose, attitude, and velocity components evolve as
\begin{align}
p_t^x&=p_{t-1}^x+\Delta t\bigl(v_{t-1}\cos\theta_{t-1}\cos\psi_{t-1}+d_{t-1}^x\bigr),\nonumber\\
p_t^y&=p_{t-1}^y+\Delta t\bigl(v_{t-1}\cos\theta_{t-1}\sin\psi_{t-1}+d_{t-1}^y\bigr),\nonumber\\
p_t^z&=p_{t-1}^z+\Delta t\bigl(v_{t-1}\sin\theta_{t-1}+d_{t-1}^z\bigr),\nonumber\\
\psi_t&=\wrap(\psi_{t-1}+\Delta t\,\omega_{t-1}^\psi),
\qquad
\theta_t=\wrap(\theta_{t-1}+\Delta t\,\omega_{t-1}^\theta),\nonumber\\
v_t&=v_{t-1}+\rho_v\bigl(v_{\rm cmd}(t-1)-v_{t-1}\bigr)\Delta t,\nonumber\\
\omega_t^\psi&=\omega_{t-1}^\psi+\rho_\psi\bigl(\omega^\psi_{\rm cmd}(t-1)-\omega_{t-1}^\psi\bigr)\Delta t,\nonumber\\
\omega_t^\theta&=\omega_{t-1}^\theta+\rho_\theta\bigl(\omega^\theta_{\rm cmd}(t-1)-\omega_{t-1}^\theta\bigr)\Delta t,
\label{eq:adx-uro20d-dynamics-pose}
\end{align}
and the remaining 12 latent parameters persist as identity maps in $\bar{\mathbf f}$, so they evolve as Gaussian random walks once $\mathbf v_t$ is added,
\begin{equation}
\begin{gathered}
b_t^v=b_{t-1}^v,\quad
b_t^{\omega^\psi}=b_{t-1}^{\omega^\psi},\quad
b_t^{\omega^\theta}=b_{t-1}^{\omega^\theta},\\
b_t^{r_j}=b_{t-1}^{r_j}\ (j=1,\ldots,4),\quad
d_t^\xi=d_{t-1}^\xi\ (\xi\in\{x,y,z\}),\quad
s_t^v=s_{t-1}^v,\quad s_t^r=s_{t-1}^r.
\end{gathered}
\label{eq:adx-uro20d-dynamics-latent}
\end{equation}
The command schedule is
\begin{align}
v_{\rm cmd}(t)&=0.88+0.16\sin(0.26t)+0.05\cos(0.11t),\\
\omega^\psi_{\rm cmd}(t)&=0.020+0.16\sin(0.31t)+0.04\cos(0.13t),\\
\omega^\theta_{\rm cmd}(t)&=0.015+0.08\sin(0.29t)+0.03\cos(0.10t).
\end{align}
Process noise is $\mathbf v_t\sim\mathcal N(\mathbf 0,\mathbf Q)$ with $\mathbf Q=\diag(\boldsymbol\sigma_f^2)$ and per-coordinate standard deviations
\begin{equation}
\begin{aligned}
\boldsymbol\sigma_f=\big[\;
&0.070,\,0.070,\,0.060,\;0.8^\circ,\,0.7^\circ,\;0.070,\,0.9^\circ,\,0.8^\circ,\;0.006,\,0.10^\circ,\\
&0.10^\circ,\;0.008,\,0.008,\,0.008,\,0.008,\;0.010,\,0.010,\,0.010,\;0.006,\,0.003
\;\big]^\top,
\end{aligned}
\label{eq:adx-uro20d-process-noise}
\end{equation}
ordered as in Eq.~\eqref{eq:adx-uro20d-state}. The raw biased and scale-corrected ranges are
\begin{equation}
\mathbf r_t(\mathbf x_t)=\bigl[\,s_t^r\rho_{1,t}+b_t^{r_1},\;
s_t^r\rho_{2,t}+b_t^{r_2},\;
s_t^r\rho_{3,t}+b_t^{r_3},\;
s_t^r\rho_{4,t}+b_t^{r_4}\,\bigr]^\top,
\quad
\rho_{j,t}=\|\mathbf p_t-\mathbf a_j\|_2.
\label{eq:adx-uro20d-raw-ranges}
\end{equation}
The proprioceptive observation channels are
\begin{equation}
\mathbf d_t(\mathbf x_t)=
\bigl[\,s_t^v v_t+b_t^v,\;
\omega_t^\psi+b_t^{\omega^\psi},\;
\omega_t^\theta+b_t^{\omega^\theta}\,\bigr]^\top.
\end{equation}
The current stored observation has four unordered range returns and three labeled proprioceptive measurements. Equivalently, with independent range noise $\boldsymbol\epsilon_t^r\sim\mathcal N(\mathbf 0,0.055^2\mathbf I_4)$ and proprioceptive noise $\boldsymbol\epsilon_t^d\sim\mathcal N(\mathbf 0,\mathbf R_d)$,
\begin{equation}
\mathbf y_t=
\begin{bmatrix}
\operatorname{sort}\!\bigl(\mathbf r_t(\mathbf x_t)+\boldsymbol\epsilon_t^r\bigr)\\
\mathbf d_t(\mathbf x_t)+\boldsymbol\epsilon_t^d
\end{bmatrix}
\in\mathbb R^7,
\qquad
\mathbf R_d=\diag\!\bigl(0.045^2,\,(1^\circ)^2,\,(1^\circ)^2\bigr).
\label{eq:adx-uro20d-observation}
\end{equation}
Because sorting destroys the anchor-to-range correspondence, the range likelihood is permutation invariant. Let $\mathbf y_t=[(\mathbf y_t^r)^\top,(\mathbf y_t^d)^\top]^\top$ with $\mathbf y_t^r\in\mathbb R^4$ and $\mathbf y_t^d\in\mathbb R^3$. Let $\mathcal S_4$ denote the symmetric group of all $4!=24$ permutations of $\{1,2,3,4\}$, and let $\mathbf P_\pi\in\{0,1\}^{4\times 4}$ denote the permutation matrix associated with $\pi\in\mathcal S_4$. The likelihood used is
\begin{equation}
p(\mathbf y_t\mid\mathbf x_t)
=
\left[
\frac{1}{|\mathcal S_4|}
\sum_{\pi\in\mathcal S_4}
\mathcal N\!\bigl(\mathbf y_t^r;\;\mathbf P_\pi\,\mathbf r_t(\mathbf x_t),\;0.055^2\,\mathbf I_4\bigr)
\right]
\mathcal N\!\bigl(\mathbf y_t^d;\;\mathbf d_t(\mathbf x_t),\;\mathbf R_d\bigr),
\label{eq:adx-uro20d-likelihood}
\end{equation}
evaluated numerically via $\operatorname{logsumexp}$ over the 24 assignments. The filtering prior, truth-generation prior, and proposal-training initial distribution are the same two-component positive-$z$ Gaussian mixture,
\begin{equation}
k\sim\Cat(0.5,0.5),\qquad
\mathbf x_0\,|\,k\sim\mathcal N\!\bigl(\boldsymbol\mu_0^{(k)},\boldsymbol\Sigma_0^{\mathrm{mode}}\bigr),
\qquad
\mathbf x_0^\star\sim p_0(\mathbf x_0),
\label{eq:adx-uro20d-filtering-prior-cat}
\end{equation}
with means
\begin{equation}
\begin{aligned}
\boldsymbol\mu_0^{(1)}=\big[\,& 0.10,\,-0.75,\,1.55,\,0.14,\,0.18,\,0.92,\,0.08,\,0.04,\,0,\,0,\\
& 0,\,0,\,0,\,0,\,0,\,0,\,0,\,0,\,1,\,1\,\big]^\top,\\
\boldsymbol\mu_0^{(2)}=\big[\,& 0.65,\,-0.30,\,2.65,\,0.14,\,0.18,\,0.92,\,0.08,\,0.04,\,0,\,0,\\
& 0,\,0,\,0,\,0,\,0,\,0,\,0,\,0,\,1,\,1\,\big]^\top,
\end{aligned}
\label{eq:adx-uro20d-filtering-prior-means}
\end{equation}
and covariance
\begin{equation}
\begin{aligned}
\boldsymbol\Sigma_0^{\mathrm{mode}}=\diag\big(\;
&0.35^2,\,0.32^2,\,0.14^2,\;(8^\circ)^2,\,(8^\circ)^2,\;0.16^2,\,(2.4^\circ)^2,\,(2.2^\circ)^2,\;0.06^2,\,(0.8^\circ)^2,\\
&(0.8^\circ)^2,\;0.08^2,\,0.08^2,\,0.08^2,\,0.08^2,\;0.06^2,\,0.06^2,\,0.06^2,\;0.03^2,\,0.020^2
\;\big),
\end{aligned}
\label{eq:adx-uro20d-filtering-prior-cov}
\end{equation}
ordered as in Eq.~\eqref{eq:adx-uro20d-state}.

\end{document}